\documentclass[final]{siamltex}
\usepackage{graphicx,color}
\usepackage{diagbox}
\usepackage{multirow}

\usepackage{epstopdf}
\usepackage{comment}
\usepackage{braket}
\usepackage{caption}
\usepackage{adjustbox}
\usepackage{amsmath, amssymb, amsfonts}
\usepackage[subrefformat=parens,labelformat=parens]{subfig}
\usepackage{float}
\usepackage{lipsum}
\usepackage{xcolor}
\usepackage{bbm}
\usepackage{enumerate}
\usepackage{bm}
\usepackage{mwe}
\usepackage{array}

\usepackage[linesnumbered,ruled,vlined]{algorithm2e}
\usepackage{algpseudocode}

\SetCommentSty{mycommfont}
\SetKwInput{KwInput}{Input}                % Set the Input
\SetKwInput{KwOutput}{Output}              % set the Output

\algrenewcommand\textproc{}

\newcommand{\Tr}{\text{Tr}}

\newcommand{\abs}[1]{\lvert#1\rvert}
\newcommand{\norm}[1]{\lVert#1\rVert}

\newcommand{\rr}{\mathbb{R}}

\newcommand{\nn}{\mathbb{N}}

\usepackage[final]{pdfpages}

\newcommand{\iprod}[1]{\langle #1 \rangle}

\newcommand{\argmin}{\operatornamewithlimits{argmin}}

\graphicspath{{./fig/}}
\makeatletter
\def\BState{\State\hskip-\ALG@thistlm}
\makeatother
\global\long\def\R{\mathbb{R}}

\global\long\def\bx{\mathbf{x}}
\global\long\def\bz{\mathbf{z}}
\global\long\def\bt{{\boldsymbol{\theta}}}

\global\long\def\Tr{\text{Tr}}

\numberwithin{equation}{section}
\numberwithin{figure}{section}

\providecommand{\corollaryname}{Corollary}
\providecommand{\lemmaname}{Lemma}
\providecommand{\propositionname}{Proposition}
\providecommand{\remarkname}{Remark}
\providecommand{\theoremname}{Theorem}

\newcommand{\rowname}[1]% #1 = text
{\rotatebox{90}{\makebox[\tempdima][c]{\textbf{#1}}}}
\allowdisplaybreaks

\title{Tensorizing flows: a tool for variational inference}

\author{
Yuehaw Khoo\thanks{Department of Statistics, University of Chicago, Illinois, IL 60637, USA. Email: {\tt ykhoo@uchicago.edu}}
\and
Michael Lindsey\thanks{Department of Mathematics, University of California, Berkeley, Berkeley, CA 94720, USA.  Email: \texttt{lindsey@math.berkeley.edu}}
\and
Hongli Zhao\thanks{Committee on Computational and Applied Mathematics, University of Chicago, Illinois, IL 60637, USA. Email: {\tt honglizhaobob@uchicago.edu}}
}

\begin{document}
\maketitle
\begin{abstract}
    Fueled by the expressive power of deep neural networks, normalizing flows have achieved spectacular success in generative modeling, or learning to draw new samples from a distribution given a finite dataset of training samples. Normalizing flows have also been applied successfully to variational inference, wherein one attempts to learn a sampler based on an expression for the log-likelihood or energy function of the distribution, rather than on data. In variational inference, the unimodality of the reference Gaussian distribution used within the normalizing flow can cause difficulties in learning multimodal distributions. We introduce an extension of normalizing flows in which the Gaussian reference is replaced with a reference distribution that is constructed via a tensor network, specifically a matrix product state or tensor train. We show that by combining flows with tensor networks on difficult variational inference tasks, we can improve on the results obtained by using either tool without the other.
\end{abstract}

% REQUIRED

% REQUIRED

\section{Introduction} 
Normalizing flows \cite{TabakVE,TabakTurner} define a category of
probability distributions from which independent samples can be drawn
directly. The deployment of neural network parametrizations \cite{RezendeMohamed}
has supercharged the practical expressivity of normalizing flows,
yielding noteworthy successes in generative modeling, i.e., the task
of learning to sample from a distribution given limited examples. Normalizing flows are based on the idea of transforming a Gaussian base distribution via a sequence of invertible mappings, and this basic concept has been realized through an increasingly diverse array of architectures~\cite{nice,https://doi.org/10.48550/arxiv.1605.08803,https://doi.org/10.48550/arxiv.1807.03039,https://doi.org/10.48550/arxiv.1806.07366}. For a recent
review of normalizing flows and their myriad applications, see \cite{FlowReview}.

Among the other key paradigms for modern generative modeling, notably
generative adversarial networks (GANs) \cite{https://doi.org/10.48550/arxiv.1406.2661} and variational
autoencoders \cite{https://doi.org/10.48550/arxiv.1312.6114}, normalizing flows are distinguished
by the fact that the samples from the model come equipped with exact sample densities.
This feature enables the direct application of normalizing flows to
variational inference, e.g., in \cite{doi:10.1126/science.aaw1147,Albergo_2019}, where one seeks to learn to sample
from or compute expectations with respect to a probability distribution whose density function $p(\bf{x})$ is known \emph{a
priori} (up to a normalization constant).

Variational inference (VI)~ \cite{veryoldvipaper,Blei_2017} has
emerged in recent decades as an alternative to Markov chain Monte
Carlo (MCMC) methods \cite{https://doi.org/10.48550/arxiv.1909.12313}, which can be confounded by long autocorrelation
times despite in-principle exactness. In spite of the contrast, VI can be combined effectively with MCMC, as in \cite{Gabri__2022,https://doi.org/10.48550/arxiv.2209.15571}. VI has found diverse applications from image processing and generation \cite{NIPS2014_d523773c,https://doi.org/10.48550/arxiv.1902.01950}, to reinforcement learning \cite{https://doi.org/10.48550/arxiv.1811.01132}, to condensed matter physics \cite{PhysRevE.84.046706,chemistryvi}, and beyond.

VI starts with a parametric family $\{ p_{\boldsymbol{\theta}} \}$ of probability densities and seeks the best approximation  of the target density $p$ lying within this family by minimizing the Kullback-Leibler (KL) divergence 
\[
D_{\mathrm{KL}}(p_{\boldsymbol{\theta}} \Vert p) = \int \log \left( \frac{p_{\boldsymbol{\theta}}(\bf{x})}{p(\bf{x})} \right) p_{\boldsymbol{\theta}}(\bf{x})  \,d \bf{x}
\]
over the low-dimensional vector of variational parameters $\boldsymbol{\theta}$. One major difficulty of VI is designing parametric classes of probability distributions that are sufficiently expressive while still permitting exact sampling with efficient density evaluation. Outside of normalizing flows, traditional approaches to VI have included parametrization via mixture models \cite{https://doi.org/10.48550/arxiv.1206.4665}, inference networks \cite{https://doi.org/10.48550/arxiv.1402.0030}, and implicit VI \cite{https://doi.org/10.48550/arxiv.1708.01529}. In flow-based approaches to VI, $p_{\boldsymbol{\theta}}$ is parametrized as the pushforward of a base Gaussian distribution $p_0$ by a map $T_{\boldsymbol{\theta}}$ which permits efficient inversion and Jacobian computation.

The other key difficulty of VI is optimizing the model parameters themselves, especially in the case where the target distribution $p$ is multimodal. In the
most extreme scenario, if the target is multimodal with well-separated modes and one's current variational guess $p_{\boldsymbol{\theta}}$ captures just a single modal component accurately, local optimization of the variational parameters cannot see the missing modes, and
they are never successfully learned. In applications of normalizing flows to VI, the unimodality of the Gaussian reference $p_0$ results in a tendency toward ``mode collapse'' during parameter optimization, wherein only a single mode is learned accurately and others remain undiscovered. Even if the reference could be chosen, e.g., as a mixture of Gaussians, there is no guarantee that all modes will
be discovered without any \emph{a priori }knowledge about the modal
structure. See \cite{GabrieBayesian} for an overview
of these difficulties.

In principle, this problem would be cured if it were possible to instead
optimize the KL divergence with the roles of $p_{\boldsymbol{\theta}}$ and $p$
exchanged, i.e., $D_{\mathrm{KL}}(p\Vert p_{\boldsymbol{\theta}})$, which is in
fact the objective used in generative modeling \cite{RezendeMohamed}.
However, to train this objective, one must be able to sample from
$p$, which is precisely the task that we are trying to accomplish
in the first place. In \cite{GabrieBayesian, https://doi.org/10.48550/arxiv.2209.15571},
schemes are proposed that use MCMC to adaptively sample from an increasingly
accurate approximation for $p$ as $p_{\boldsymbol{\theta}}$ is trained. However,
although such approaches can avoid mode collapse, they cannot fundamentally
extend the capacity of the training to discover new modes (except
those that can be discovered by reasonably long MCMC exploration).

More fundamentally, all uses of normalizing flows rely heavily on
the expressive power of neural networks to correct the base distribution
$p_{0}$. If $p$ is far from Gaussian, even if an approximation $p_{\boldsymbol{\theta}} \approx p$
can be effectively learned, a good approximation may require a
neural network parametrization that is expensive to train.

Therefore, there is ample motivation to construct a reference $p_0$
that approximates the global structure of $p$. If this can be accomplished,
one might bypass difficulties imposed by multimodality and moreover,
get away with more compact and easily trainable neural network parametrizations of the flow map. Several works have considered improvements of the base distribution. For capturing heavy-tailed distributions, \cite{https://doi.org/10.48550/arxiv.1907.04481} has considered the use of density quantile function, whereas \cite{https://doi.org/10.48550/arxiv.2107.07352} proposes to use copulas as base distributions. Meanwhile, \cite{https://doi.org/10.48550/arxiv.2110.15828} adaptively exploits accept-reject re-sampling. However, none of these approaches can incorporate \emph{a priori} knowledge about the global modal structure of the target distribution.

Tensor networks originally emerged from the numerical study of quantum many-body problems. They provide low-dimensional parametrizations capable of overcoming the curse of dimensionality that prevents many-particle wavefunctions from being stored exactly. The most widely used and successful tensor network remains the matrix product state (MPS), which emerges in tandem with the density matrix renormalization group (DMRG) algorithm for computing many-body ground states \cite{PhysRevB.48.10345}. The same structure has also appeared as the tensor train (TT) \cite{doi:10.1137/090752286} in the applied mathematics literature, where it is often viewed within the broader context of generalizing the singular value decomposition to higher-order tensors \cite{doi:10.1137/07070111X, doi:10.1137/S0895479896305696}. In recent years, low-rank tensor decompositions have received increasing attention, for example, in surrogate modeling and deep learning \cite{Or_s_2014,Cichocki_2016, https://doi.org/10.48550/arxiv.1605.05775}. Among many formats, the TT format is distinguished by the ease with which it allows for various key operations to be performed, with complexity that depends only linearly on the problem dimension, as long as the core tensor ranks remain bounded. Moreover, various TT operations also support convenient parallelism \cite{https://doi.org/10.48550/arxiv.2111.10448,https://doi.org/10.48550/arxiv.2011.06532}. Many high-dimensional classical problems, such as generative modeling \cite{https://doi.org/10.48550/arxiv.2202.11788}, variational Bayes inference \cite{https://doi.org/10.48550/arxiv.2010.06564, 8903177}, supervised learning \cite{https://doi.org/10.48550/arxiv.2101.09184}, numerical partial differential equations \cite{doi:10.1137/100785715, https://doi.org/10.48550/arxiv.2207.01962}, and dynamical systems \cite{Klus_2018} can in important instances be recast and solved efficiently in the TT format.

In a VI setting, one is given (up to normalization) the density $p(\bx)$ as a black-box function. Therefore, cross-approximation can be applied to obtain the components of a TT. In this work, we propose to combine the advantages of NF and TT for the purpose of VI. We construct a base TT distribution by squaring a functional tensor train \cite{Soley_2021} obtained from a cross approximation  \cite{OSELEDETS201070} of the square root of the target density \cite{pmlr-v161-novikov21a}. Relative to \cite{https://doi.org/10.48550/arxiv.1810.01212}, this ensures the non-negativity and exact sample-ability of the base distribution. Then we push forward our base distribution by a flow map parametrized with neural networks. We refer to our model as a \emph{tensorizing flow} (TF) since the advantages of both tensor decompositions and flow-based models are combined. Like NF, TF admits exact sampling equipped with sample densities. However, it demonstrates improved performance over straightforward NFs at comparable computational cost, measured in terms of the variational objective. Meanwhile, the flow is able to improve the base TT distribution obtained directly with numerical linear algebra routines. We demonstrate the advantage of our method on a Ginzburg-Landau model~\cite{Hohenberg_2015, Li_2019} that we find to be intractable to approach with either NF or TT individually, as well as other high-dimensional test distributions of interest.

\textbf{Our contribution:} In this paper, we tackle the problem of initializing the base distribution for a normalizing flow. This is achieved with tensor network methods, which ultimately reduce to fast linear algebra routines.  In particular, we leverage tensor-train orthogonalizations to improve the linear scaling in the recently proposed squared inverse Rosenblatt transport \cite{inverse-rosenblatt-tt} method for interpolating probability densities with guaranteed non-negativity. From another perspective, our method can be viewed as an improvement over TT-based generative model \cite{squared-inverse-rosenblatt}, by augmenting it with a neural-network flow model, as TT has a limited representation power. 

\subsection{Organization} We review the background materials on VI and NF in Section~\ref{background-and-prelim}. In Section~\ref{proposed-method} we detail the  methodology for constructing the base distribution using a squared TT representation and for generating samples (equipped with densities) at computational cost scaling linearly with dimension for fixed ranks. The final TF model is tested on a variety of high-dimensional probability distributions in  Section~\ref{numerical-experiments}.

\subsection{Notation} \label{common-notations} For a positive integer $n\in \nn$, throughout we denote $[n] := \{1,2,\ldots, n\}$. Throughout the text, $d$ shall denote the problem dimension, i.e., the dimension of the sampling variable $\bx$. Given $n_1,\ldots, n_k\in\nn$ and indices $i_j\in [n_j]$, $1\le j\le d$, we will often consider multi-indices of the form $\mathcal{I}:=(i_1,\cdots, i_d)$.
We use square brackets to represent discretely indexed tensor values, e.g., $\mathcal{A}[\mathcal{I}]$ for a suitable $d$-tensor $\mathcal{A}$. Meanwhile we use parentheses to denote continuous function evaluations, e.g., $p(\mathbf{x})$. To describe sub-tensors of a tensor, we adopt the widely-used `MATLAB notation.' For instance, if $\mathbf{A}\in\rr^{m\times n}$ denotes a matrix or 2-tensor, then $\mathbf{A}[i,j]$ represents a single element, while $\mathbf{A}[i,:]$ and $\mathbf{A}[:,j]$ represent the $i$-th row  and  $j$-th column, respectively. In naming objects, we will use non-boldface lowercase letters to denote scalars (e.g., $a, u, v\in\rr$), boldface lowercase letters for vectors (e.g., $\mathbf{v}, \mathbf{w}\in\rr^n$), boldface uppercase letters for matrices (e.g., $\mathbf{A}, \mathbf{B}\in\rr^{m\times n}$), and calligraphic uppercase letters for general high-dimensional tensors (e.g., $\mathcal{A}, \mathcal{C}\in\rr^{n_1\times\cdots\times n_d}$). There may be exceptions to this rule as becomes necessary, but these will be explicitly noted. 

Finally, several tensor network diagrams \cite{Bridgeman_2017} are presented to better illustrate tensor train operations. In this work, a tensor diagram is composed of nodes and edges, where the nodes represent a `core tensors' out of which a large tensor associated to the diagram is built. In general, we allow for tensors with mixed discrete and continuous indices by thinking of them simply as functions jointly of some discrete and some continuous variables. The number of outgoing edges from a node represents the order of the tensor, i.e., the number of indices. Connections between outgoing edges indicate that the corresponding indices are shared and summed (or integrated) out in the overall tensor, cf. Section~\ref{review-tt}, where we shall review other special notations as needed.

\section{Background and preliminaries}\label{background-and-prelim} We introduce the general variational inference problem in Section~\ref{review-problemstatement} along with its implementation using normalizing flows in Section~\ref{vinf}. Section~\ref{review-tt} reviews the matrix product state (MPS) / tensor train (TT) decomposition, summarizes necessary operations in this format, and discusses its application to efficient compression of high-dimensional probability distributions.
\label{sec:background}
\subsection{Variational inference}\label{review-problemstatement}
Given a function $U:\Omega\rightarrow\rr$, we would like to learn a target density defined as:
\begin{equation}\label{eq:1}
    p(\mathbf{x}) = \frac{1}{Z} e^{-U(\mathbf{x})},
\end{equation} where $Z = \int_{\Omega} e^{-U(\mathbf{x})}\,d\mathbf{x}$ is the normalizing constant or partition function.

In variational inference (VI), one formulates the partition function in terms of the optimal value of an optimization problem. More precisely, given a family of distributions $\mathcal{P}$, one minimizes 
\begin{eqnarray}\label{exact-flow-loss}
  L(\tilde p )&:=&  \mathbb{E}_{\mathbf{x}\sim \tilde{p}}\bigg[
        \log \tilde {p}(\mathbf{x}) + U(\mathbf{x}) \bigg] \cr
        &=& D_{\text{KL}}(\tilde p || p) - \log Z.
\end{eqnarray}
over $\tilde p\in \mathcal{P}$. Here $D_{\text{KL}}$ denotes the Kullback-Leibler (KL) divergence.

Assuming that $p \in \mathcal{P}$, the optimal value of this minimization problem is $ -\log Z$, which is attained if and only if $D_{\text{KL}}(\tilde p || p) = 0$, i.e., if and only if $\tilde{p} = p$. If $p \notin \mathcal{P}$, then optimizing \eqref{exact-flow-loss} recovers the best approximation of $p$ within the family $\mathcal{P}$, as measured by the KL divergence.

The expectation in (\ref{exact-flow-loss}) can be estimated using a finite set of samples $\{\mathbf{x}^{(s)}\}^{S}_{s=1}$ drawn from $\tilde p$, provided we can compute the densities $\tilde{p}(\bx^{(s)})$ of our samples.

\subsection{Variational inference with normalizing flows}\label{vinf} In practical applications, we consider a family of distributions parametrized by $\boldsymbol{\theta}$. Denoting now the parametrized distribution as ${p}_{\boldsymbol{\theta}}$, we seek to solve:
$$
    \boldsymbol{\theta}^\star = \argmin_{\boldsymbol{\theta}}L(\bt),
$$
where we now abuse notation slightly be identifying $L(\bt) = L(p_\bt)$.

The accuracy of VI is ultimately limited by the expressivity of the parametrized family $\{p_{\boldsymbol{\theta}}\}$.
In this work, we focus on families based on normalizing flows. More precisely, we construct ${p}_{\boldsymbol{\theta}}$ as the \emph{pushforward} of a fixed base distribution $p_0$ by the composition of a sequence $F^{(k)}_\bt$, $k=1,\ldots,K $, of differentiable invertible maps, as in
\[
T_\bt := F^{(K)}_\bt \circ  \cdots \circ F^{(1)}_\bt,
\]
which are themselves parametrized by a collection of neural network parameters $\bt$. The number $K$ of maps is the length of the flow. It is useful to denote 
\[
T_\bt^{(k)} := F_\bt^{(k)} \circ \cdots \circ F_\bt^{(1)},
\]
with $T_\bt^{(0)} := \mathrm{Id}$, 
so $T_\bt^{(k+1)}  = F_\bt^{(k+1)} \circ  T_\bt^{(k)}$ for $k=0,\ldots, K-1$, and $T_\bt^{(K)} = {T}_\bt$.

Concretely, to sample $\bx \sim p_\bt$, one samples $\mathbf{z} \sim p_{0}$ and computes 
$\mathbf{x} = T_{\boldsymbol{\theta}}(\mathbf{z})$. 
The model density of a sample $\bx$ can be computed by change of variables as 
\[
p_\bt (\bx) = p_0(T_\bt^{-1}(\bx))\,\vert DT_\bt^{-1} (\bx) \vert.
\]
Then observe that the loss can be written as an expectation with respect to the base distribution:
\[
L(\bt) = \mathbb{E}_{\bz \sim p_0} \Big[ \log p_0(\bz) - \log \vert DT_\bt (\bz) \vert  + U(T_\bt (\bz)) \Big],
\]
which can be further unpacked as 
\begin{equation} \label{eq:Ltheta_unpack}
    L(\bt) = \mathbb{E}_{\bz \sim p_0} \left[ \log p_0(\bz) - \sum_{k=1}^K \log \left\vert DF_\bt^{(k)} \left( T^{(k-1)}_\bt (\bz) \right) \right\vert  + U(T_\bt (\bz)) \right].
\end{equation}

By drawing $S$ independent samples $\bz^{(1)}, \ldots , \bz^{(S)} \sim p_0$, the loss \eqref{eq:Ltheta_unpack} can be estimated empirically as 
\begin{equation} \label{eq:Lhattheta_unpack}
    \widehat{L}(\bt) = \frac{1}{S} \sum_{s=1}^S \left[ \log p_0(\bz^{(s)}) - \sum_{k=1}^K \log \left\vert DF_\bt^{(k)} \left( T^{(k-1)}_\bt (\bz^{(s)}) \right) \right\vert  + U(T_\bt (\bz^{(s)})) \right],
\end{equation}
and the gradient $\nabla_\bt L (\bt)$ can be estimated via the `reparametrization trick,' i.e., by freezing the samples $\bz^{(1)}, \ldots , \bz^{(S)}$ and automatically differentiating through the evaluation formula \eqref{eq:Lhattheta_unpack}.

The sequence of mappings $F_{\boldsymbol{\theta}}^{(1)},\cdots, F_{\boldsymbol{\theta}}^{(K)}$ are chosen to balance expressitivity with efficient computation of the determinant of the Jacobian. We refer the reader to \cite{Kobyzev_2021} for a variety of choices. Some popular architectures include NICE \cite{nice}, FFJORD \cite{https://doi.org/10.48550/arxiv.1810.01367}, and RealNVP \cite{https://doi.org/10.48550/arxiv.1605.08803}.
In this work we opt for invertible residual networks, or iResNets~\cite{https://doi.org/10.48550/arxiv.1811.00995,https://doi.org/10.48550/arxiv.1906.02735}.

\subsection{Tensor operations}\label{review-tt} In this section, we summarize certain fundamental tensor operations. 

We illustrate tensor contraction and the associated notation with illustrative examples. More formal definitions can be found, for example, in \cite{https://doi.org/10.48550/arxiv.2109.00626}. Tensor contraction is loosely defined as summation/integration over repeated indices/arguments. For instance, consider a 3-tensor $\mathcal{A}\in\rr^{n_1\times n_2\times n_3}$ and a matrix $\mathbf{B}\in\rr^{n_2\times m_2}$. A new tensor $\mathcal{C}$ of size $n_1\times m_2\times n_3$ can be formed by contracting the middle index of $\mathcal{A}$ with the first index of $\mathbf{B}$:
\begin{equation}\label{discrete-contraction}
    \mathcal{C}[i_1,j_2,i_3] = \sum_{k_2}\mathcal{A}[i_1,k_2,i_3]\mathbf{B}[k_2,j_2] =: \iprod{\mathcal{A}[i_1, k_2, i_3], \mathbf{B}[k_2, j_2]}_{k_2\in [n_2]}
\end{equation}
One may consider a function to be a tensor by identifying indices with function arguments, and one can define suitable tensor contractions analogously. For example, letting $k:X_1\times X_2\rightarrow\rr$ and $g:X_2\rightarrow\rr$, a new function $f:\rr\rightarrow\rr$ can be defined as:
\begin{equation}\label{continuous-contraction}
    f(x_1) = \int_{X_2}k(x_1,x_2)g(x_2)dx_2 = \iprod{k(x_1, x_2), g(x_2)}_{x_2\in X_2}.
\end{equation} One can also define a mixed-type contraction such as the following\footnote[1]{When there are mixed (i.e., both discrete and continuous) indices, we generally prefer to write the discrete indices as subscripts.}:
\begin{equation}\label{mixed-type-contraction}
    \mathcal{C}_{i_1i_3}(x_2) := \mathcal{C}[i_1,x_2,i_3] = \sum_{i_2}\mathcal{C}[i_1,i_2,i_3]f_{i_2}(x_2) =: \iprod{\mathcal{C}[i_1,i_2,i_3], f_{i_2}(x_2)}_{i_2\in [n_2]}
\end{equation} 

In the above equations, we introduced the bracket notation as a shorthand to denote contractions over certain indices / continuous spaces. We will occasionally revisit this notation to render lengthy summations over multiple indices more readable. 

In this work, special cases of contractions such as tensor-tensor, tensor-matrix and tensor-vector contractions are frequently encountered. Contractions can be used to combine low-order tensors into a high-order tensor. For instance, a tensor train constructs a $d$ dimensional tensor from several $3$-tensors.

\begin{figure}[ht]
\centering
\setkeys{Gin}{draft=false}
\includegraphics[width = 3.5in]{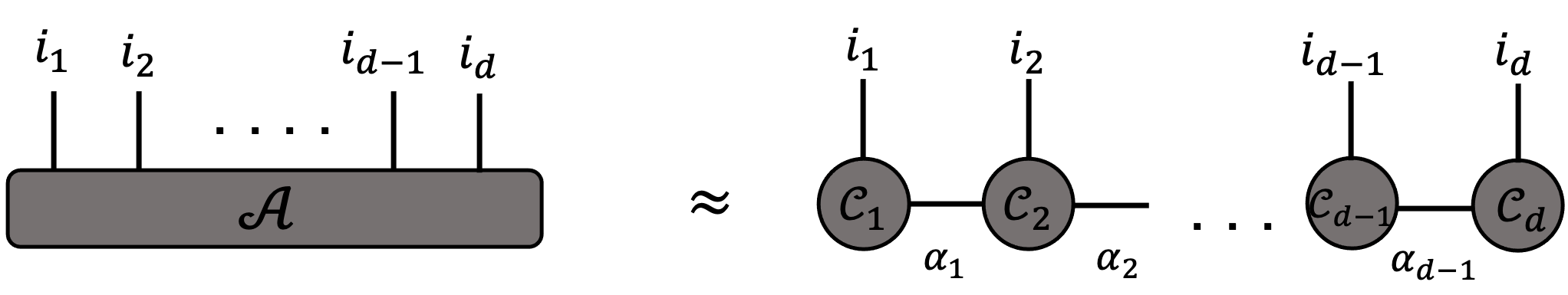}
\caption{Tensor-train decomposition of a $d$-dimensional tensor }\label{discrete-tt}
\end{figure}

\begin{definition}\label{define-tensortrain} (Tensor train decomposition) A tensor $\mathcal{A}\in\mathbb{R}^{n_1\times\cdots \times n_d}$ is a tensor train (TT)  with ranks $(r_0,\cdots, r_{d})$, where $r_0=r_d=1$, if it can be written
\begin{eqnarray*}\label{discrete-tt-format}
    \mathcal{A}[\mathcal{I}] &=& \sum_{\alpha_1=1}^{r_1}\cdots\sum_{\alpha_{d-1}=1}^{r_{d-1}}\mathcal{C}_1[1,i_1,\alpha_1]\mathcal{C}_2[\alpha_1,i_2,\alpha_2]\cdots\mathcal{C}_d[\alpha_{d-1},i_d,1]\\
   &=& \mathcal{C}_1[1,i_1,:] \mathcal{C}_2[:,i_2,:]\cdots \mathcal{C}_d[:,i_d,1],
\end{eqnarray*}
where $\mathcal{I} = (i_1,\ldots,i_d)$, the last equality is simply a sequence of matrix multiplications, and the 3-tensors $\mathcal{C}_i\in\mathbb{R}^{r_{i-1}\times n_i\times r_i}$ are called the core tensors. 
\end{definition} 

A tensor diagram illustration is presented in Figure~\ref{discrete-tt}, using conventions described in Section~\ref{common-notations}.

The advantage of this decomposition is that it allows us to store and compute with the list of cores $\{\mathcal{C}_i\}_{i=1}^{d}$, instead of the full tensor $\mathcal{A}$, which has $O(n^d)$ elements. If the ranks are low, the computational cost of various operations can be reduced significantly. Indeed, storage scales as $O(dnr^2)$, and evaluation of a single entry of the tensor scales as $O(dr^2)$.  Common arithmetic operations with TTs, such as addition, can be implemented within the TT format by manipulating the cores only, as summarized in \cite{doi:10.1137/090752286}. The TT ranks can be truncated optimally via SVDs of unfolding matrices \cite{doi:10.1137/090752286}.

\begin{figure}[ht]
\centering
\setkeys{Gin}{draft=false}
\includegraphics[width = 2.5in]{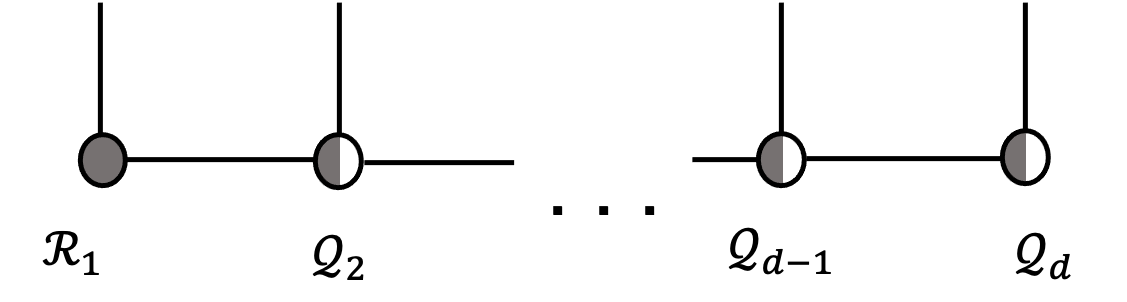}
\caption{Illustration of the TT right-left orthogonal form.}\label{fig:rightleftqrform}
\end{figure}

Finally, we introduce the orthogonalization operation on a TT, also known as canonicalization in the MPS literature. This procedure orthogonalizes each TT core in a sequential fashion until all but one core is orthogonalized. Using tensor diagram notation, we present the orthogonalized TT decomposition in Figure~\ref{fig:rightleftqrform} and refer the interested reader to~\cite{doi:10.1137/090752286,Gelss2017} for more details. We mark the orthogonalized cores as half-solid to distinguish them from the original ones, as introduced in \cite{doi:10.1137/100818893}. The light side of the node indicates that for orthogonalized cores $\mathcal{Q}_k$ have orthonormal rows after suitable matricization, i.e., that 
\begin{equation}\label{rithtleftortho-identity}
    \iprod{\mathcal{Q}_k[:,i_k,\alpha_k], \mathcal{Q}_k[:,i_k,\alpha_k]}_{i_k, \alpha_k} = I_{r_{k-1}}, \text{ for all $k=2,\ldots, d$}
\end{equation} 
The TT in Figure~\ref{fig:rightleftqrform} is referred to as \emph{right-left orthogonal}. The orthogonalization procedure costs $O(dnr^3)$ operations. 

\section{Proposed method}\label{proposed-method}
We present in this section the construction of the tensorizing flow. We consider a target distribution $p$ of form (\ref{eq:1}) to be approximated on a domain $\Omega = [a_1,b_1]\times\cdots\times[a_d,b_d]\subset\rr^d$. By shifting and scaling the domain, we can always assome that $\Omega = [-1,1]^d$. Within the flow framework of \eqref{eq:Ltheta_unpack}-\eqref{eq:Lhattheta_unpack}, rather than choosing a Gaussian base distribution $p_0$, we will use a TT to construct a base distribution $p_0$ that already approximates the target $p$ reasonably well. In order to optimize the empirical loss \eqref{eq:Lhattheta_unpack}, we are required to be able to efficiently achieve the following with $p_0$:
\begin{itemize}
    \item draw samples $\bz \sim p_0$.
    \item compute sample densities $p_0 (\bz)$.
\end{itemize}

In Section~\ref{initialize-with-tt}, we will construct $p_0 \approx p$ using a TT in a way that meets these criteria. In Section~\ref{pushforwardtt}, we will present our specific choice of flow maps $F_{\boldsymbol{\theta}}^{(k)}$ based on invertible residual networks.

\subsection{Base distribution}\label{initialize-with-tt} We will construct $p_0 (\bx) $ as the pointwise square of a functional tensor train \cite{inverse-rosenblatt-tt} denoted $q_0(\bx)$. In Section~\ref{conditionalsamplingdescribe} we explain how samples equipped with densities can be drawn from such a $p_0$, and in Section~\ref{finding} we explain how $q_0$ can be prepared using only the ability to evaluate the energy function $U(\bx)$ defining the target $p(\bx) \propto \exp(-U(\bx))$.

\subsubsection{Autoregressive sampling framework}\label{conditionalsamplingdescribe} Sampling from a multivariate distribution is in general a nontrivial task. In the style of other work using MPS/TT for generative modeling \cite{bornmachinenature,https://doi.org/10.48550/arxiv.2202.11788,pmlr-v161-novikov21a}, we will draw a sample $\bx^* \sim p_0$ `autoregressively,' i.e., by writing:
\begin{equation}\label{chain-rule}
    p_0(x_1,\ldots,x_d) = p_1(x_1)p_2(x_2|x_1)\cdots p_d(x_d|x_1,\ldots,x_{d-1}),
\end{equation}
then sampling each successive coordinate $x_k^*$ from the appropriate conditional distribution, conditioned on previously sampled coordinates $x_j$, $j=1,\ldots, k-1$.

In other words, for each $k = 1,2,\cdots, d$, we would like to sample
\begin{equation}\label{sampling-process}
    x_k^* \sim p_k(x_k|\mathbf{x}_{<k}^*),
\end{equation}
where $\mathbf{x}_{<k}^* = (x_1^*, x_2^*, \ldots, x_{k-1}^*)$.
This approach converts the problem of obtaining a multivariate sample $\mathbf{x}^* = (x_1^*,\ldots, x_d^*)$ into sequence of univariate sampling problems, which we can view as essentially trivial. The difficulty is to construct these univariate distributions. By the definition of conditional probability:
\begin{equation}\label{eqn:conditional-sampling-theorem}
    p_k(x_k|\mathbf{x}_{<k}^*) = \frac{p_0(\mathbf{x}_{<k}^*,x_k)}{p_{<k}(\mathbf{x}_{<k}^*)} \propto p_{<k+1}(\mathbf{x}_{<k}^*,x_k),
\end{equation}
viewing the previously sampled coordinates as fixed. Here we have defined the marginal distributions:
\begin{equation}\label{marginaldensitynotation}
    p_{<k}(\mathbf{x}_{<k}) := \int p_0(x_1,\ldots,x_k,x_{k+1},\ldots,x_d)\, dx_{k+1}\ldots dx_d
\end{equation}
Therefore, to construct the univariate distributions that we need, we will essentially only need to marginalize $p_0$, possibly with some leading variables fixed.

\subsubsection{Representation of $p_0$}
To deal with the continuum in even a single component, we discretize by considering a truncated orthonormal basis $\phi_1, \ldots, \phi_n$ for $L^2([-1,1])$, furnished by, e.g., orthogonal polynomials. In practice we can choose a different basis size $n_j$ for each dimension $j=1,\ldots, d$, but for simplicity of the discussion we will just consider the case of uniform $n_j = n$.

For the time being, assume that we are also given a $d$-dimensional tensor train $\mathcal{C} \in \mathbb{R}^{n\times \cdots \times n}$ that specifies the coefficients of a function 
\[
q_0(\bx) \approx q(\bx) := \sqrt{p(\bx) }
\]
in the $d$-fold tensor product basis induced by $\{\phi_i \}_{i=1}^n$. Our approximate density, $p_0 \approx p$, can then be obtained as the pointwise square $p_{0}(\bx)=q_{0}(\bx)^2$.

Concretely, suppose 
\begin{equation}\label{tensor-representation}
    q_{0}(\bx) = \sum_{\mathcal{I}} \mathcal{C}[\mathcal{I}]\, \phi_{\mathcal{I}}(\bx) = \sum_{i_1=1}^{n}\cdots\sum_{i_d=1}^{n}\mathcal{C}[i_1,\cdots, i_d]\phi_{i_1}(x_1)\cdots \phi_{i_d}(x_d),
\end{equation}
where the pure tensor product functions $\phi_{\mathcal{I}}$ and coefficient tensor $\mathcal{C}[\mathcal{I}]$ satisfy
\[
\phi_{\mathcal{I}}(\bx) := \phi_{i_1} (x_1) \cdots \phi_{i_d} (x_d),
\quad 
\mathcal{C}[\mathcal{I}] = \iprod{q_0 (\mathbf{x}), \phi_{\mathcal{I}}(\mathbf{x})}_{\mathbf{x}\in\Omega}.
\]
Let the cores of $\mathcal{C}$ be denoted  $\mathcal{C}_1,\ldots,\mathcal{C}_d$. 
The details of obtaining $\mathcal{C}$ as a TT are deferred for now to Section~\ref{finding}.

We remark here that although we intend to obtain samples from $p_{0}(\bx) = q_0(\bx)^2$, we never explicitly form its coefficient tensor as a TT, as doing so would require us to work with core tensors of size $r^2 \times n \times r^2$~\cite{squared-inverse-rosenblatt}. Instead we only work directly with the core tensors of $\mathcal{C}$. In particular, we demonstrate that by putting the cores of $\mathcal{C}$ in right-left orthogonal form, the marginalizations in variables $x_{k+1},\ldots, x_d$ required in the sampling process \eqref{sampling-process} can be performed for free, and the entire cost of drawing a sample will scale only linearly with the dimension $d$.
 
As such, before discussing further details, we define a right-left orthogonalization of $\mathcal{C}$,
$$
    \mathcal{C}[i_1,\ldots, i_d] = \mathcal{R}_1[1,i_1,:]\mathcal{Q}_2[:,i_2,:]\cdots\mathcal{Q}_d[:,i_d,1],
$$
which can be computed in $O(dnr^3)$ time as reviewed in Section~\ref{review-tt}. See Figure~\ref{fig:rightleftqrform} for a diagram.

\subsubsection{Sampling algorithm}\label{sec:sampling}
\begin{figure}[ht]
    \centering
    \includegraphics[width = 2.5in]{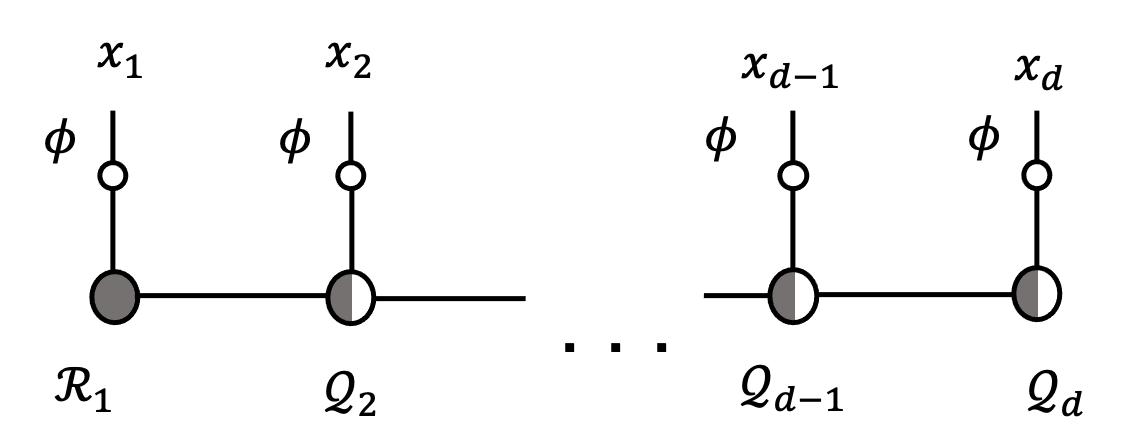}
    \caption{Illustration of $q_0(\mathbf{x})$ as a tensor network, composed of a tensor train coefficient $\mathcal{C}$ with continuous univariate functions $\phi_{i}(x_i)$ in each dimension. It can be interpreted as a mixed-type tensor with both discrete and continuous indices.  }\label{fig:functional-tt}
\end{figure}

Now we discuss how to produce exact samples $\mathbf{x}^* \sim p_0$ given $\mathcal{C}$ as a TT in right-left orthogonal form. At any step $k=1,\ldots,d$, we need to have a representation of the marginal 
\begin{equation}\label{eq:q0q0marg}
p_{< k+1}(\bx_{<k}^*,x_k) = \int q_0(x_1^*,\ldots, x_{k-1}^*,x_k,\ldots,x_d)^2 \,dx_{k+1} \cdots dx_{d} 
\end{equation}
in order to generate a coordinate sample $x_k^*$. The function $q_0(\bx)$ can be represented with a tensor network diagram with continuous external legs by attaching the mixed tensors $\phi_i(x)$ to the external legs of the TT diagram for $\mathcal{C}$, as depicted in Figure~\ref{fig:functional-tt}.

\begin{figure}[ht]
    \centering
    \includegraphics[width = 5.5in]{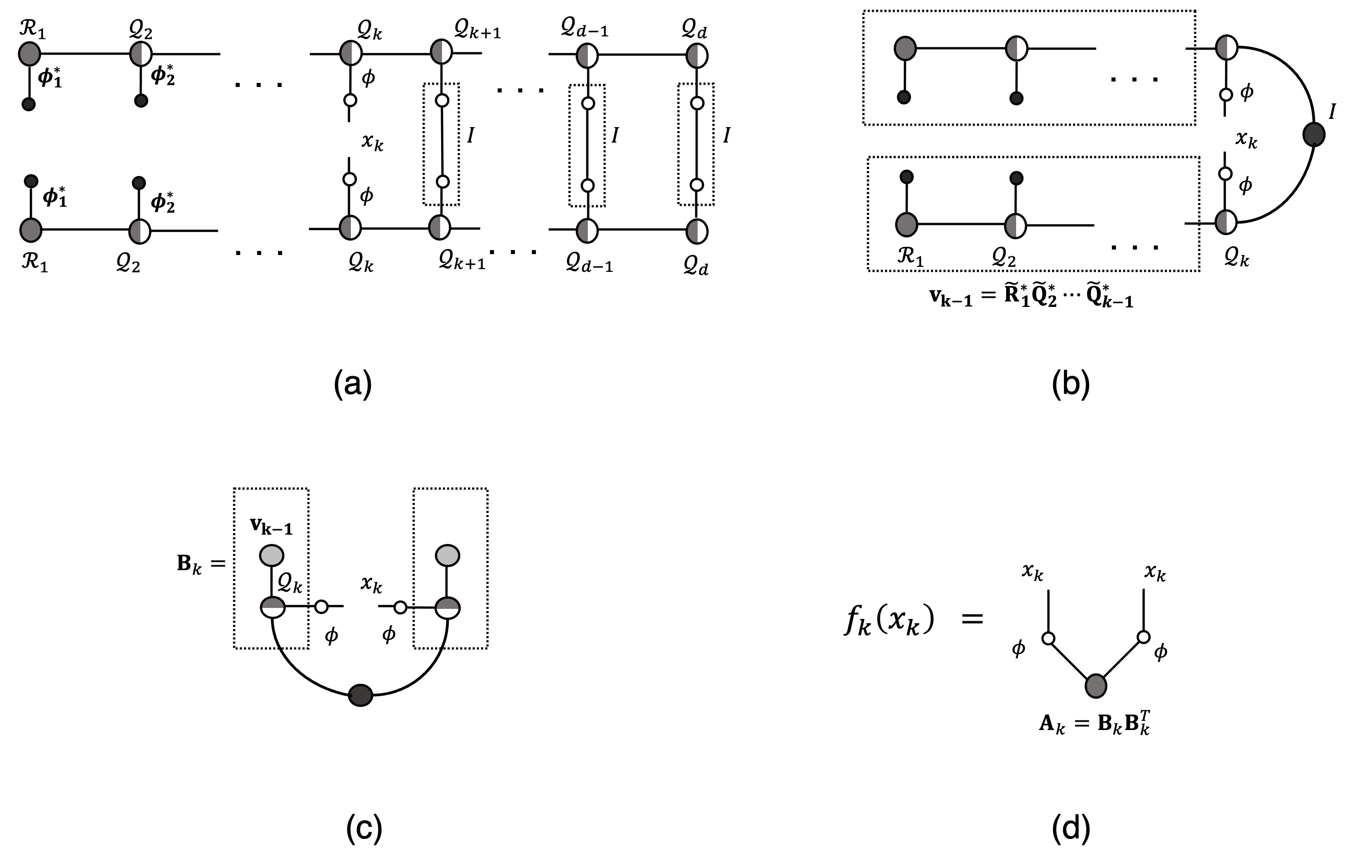}
    \caption{Construction of $f_k(x_k) := p_{<k+1}(\mathbf{x}_{<k}^*, x_k)$. For a detailed step-by-step description of the procedure, refer to the main text of Section~\ref{sec:sampling}.}
    \label{tt-conditionalsampling2}
\end{figure}

The tensor contractions required to compute \eqref{eq:q0q0marg} are depicted in Figure~\ref{tt-conditionalsampling2}. Orthonormality of the univariate basis functions $\phi_1,\ldots,\phi_n$ guarantees that the integration in \eqref{eq:q0q0marg} yields identity matrices forming the rungs of a `ladder' to the right of the $k$-th cores in Figure~\ref{tt-conditionalsampling2}(a). Meanwhile, fixing the previous coordinates $x_1^*,\ldots, x_{k-1}^*$ amounts to computing tensor-vector contractions between the cores $\mathcal{R}_1,\mathcal{Q}_2,\ldots, \mathcal{Q}_{k-1}$ and vectors $\boldsymbol{\phi}_1^*, \boldsymbol{\phi}_2^*,\ldots,\boldsymbol{\phi}_{k-1}^*$ of basis function evaluations, defined by 
\begin{equation}\label{compute-v}
        \boldsymbol{\phi}_{j}^* = 
        \begin{bmatrix}
            \phi_1(x_{j}^*)\\
            \vdots\\
            \phi_n (x_{j}^*) 
        \end{bmatrix} \in \rr^{n},
\end{equation}
as depicted in Figure~\ref{tt-conditionalsampling2}(a). There we denote the matrices resulting from these tensor-vector contractions as $\mathbf{R}_1^*,\mathbf{Q}_2^*,\ldots,\mathbf{Q}_{k-1}^*$, respectively.

In passing from frame (a) to frame (b) of Figure~\ref{tt-conditionalsampling2},
right-left orthogonality allows us to simply `collapse the ladder' from right to left in the tensor network diagram without any additional computation. To pass from frame (b) to frame (c), the left tail of the train can be collapsed into a single vector $\mathbf{v}_{k-1}$ by the indicated sequence of matrix-vector multiplications. To maintain linear dimension scaling, the computation of $\mathbf{v}_{k-2}$ can simply be recycled from the preceding step, and $\mathbf{v}_{k-1}$ can be produced with only a single additional matrix multiplication.

In frame (c), the core $\mathcal{Q}_k \in \R^{r_{k-1} \times n \times r_k}$ is contracted with $\mathbf{v}_{k-1} \in \R^{r_{k-1}}$ to produce a matrix $\mathbf{B}_k \in \rr^{n \times r_k }$. In passing from frame (c) to frame (d), we perform the matrix-matrix multiplication defining $\mathbf{A}_k := \mathbf{B}_k \mathbf{B}_k^\top$, and from frame (d) it is clear that 
\begin{equation}\label{eq:fkdef}
f_k(x_k) := p_{< k+1}(\bx_{<k}^*,x_k) = \sum_{i,j=1}^n \mathbf{A}_k[i,j] \phi_i (x_k) \phi_j(x_k).
\end{equation}
Note that $\mathbf{A}_k$ is positive semidefinite by construction, guaranteeing pointwise nonnegativity of this expression as a function of $x_k$.

Now following \eqref{eqn:conditional-sampling-theorem}, we know that the univariate function $f_k$ is proportional to the univariate conditional probability density $p_k(x_k|\mathbf{x}_{<k}^*)$ from which we wish to draw a sample. If the $\phi_1, \ldots, \phi_n$ are polynomials such as Legendre or Chebyshev polynomials, the CDF can be computed exactly and then a sample can be drawn using the inverse CDF method through efficient root-finding. In practice, it is also efficient to just evaluate $f_k$ on a finely spaced grid and implement the inverse CDF method via the trapezoidal rule.

Given a multivariate sample $\bx^* = (x_1^*,\ldots,x_d^*)$ furnished by our sampling routine, it is trivial to recover its density $p_0(\bx^*)$. Indeed, observe based on the definition~\eqref{eq:fkdef} that $f_d(x_d^*) = p_0(\bx^*)$, so we need only evaluate the final univariate function $f_d$ that we have constructed at the final coordinate sample $x_d^*$.

In summary, once $\mathcal{C}$ has been furnished as a TT, the entire cost of producing $N$ samples from $p_0$ scales as
$$
O(dnr^3 + Ndn^2r^2),
$$
Here the first term is associated to the right-left orthogonalization pre-processing step, and we have defined $r:= \max_{i=1,\ldots,d-1} \{r_i\}$ as the maximal TT rank of $\mathcal{C}$. Recall moreover that $n$ is the size of the univariate basis $\phi_1,\ldots, \phi_n$, and $d$ is the problem dimension.

\subsubsection{Obtaining $\mathcal{C}$ as a TT}\label{finding} It remains to discuss how to produce the coefficient tensor $\mathcal{C}$ of $q_0(\bx)$ in TT format.
The key difficulty of this section is to figure out how to leverage the TT-cross algorithm~\cite{OSELEDETS201070}.

Recall from \eqref{eq:1} that $p \propto \tilde{p} := e^{-U}$, where we have defined $\tilde{p}$ to be an unnormalized density that we are capable of evaluating pointwise. Recall further that we want a TT-based approximation $q_0$ of $q = \sqrt{p}$. First we will obtain a TT-based approximation $\tilde{q}_0$ of $\tilde{q} := \sqrt{\tilde{p}}$, and then we will recover an appropriate $q_0$ via a normalization step.

We will also express $\tilde{q}_0$ in terms of a coefficient tensor, namely $\mathcal{B}$, with respect to our functional tensor product basis: 
\begin{equation}\label{multivariate-basis-expansion}
    \tilde{q}_0(\bx) = \sum_{\mathcal{I}}\mathcal{B}[\mathcal{I}] \, \phi_{\mathcal{I}}(\mathbf{x}).
\end{equation}
The coefficient tensor $\mathcal{B}$ will differ from the coefficient tensor $\mathcal{C}$ that we seek by only a scalar constant factor that we will compute later.

TT-cross is most naturally suited to recovering a function defined on a multivariate grid, which can be viewed as an ordinary tensor with discrete indices. As such we will use TT-cross to recover a TT-based approximation of $\tilde{q}$ on such a grid and then use numerical quadrature to deduce an appropriate coefficient tensor $\mathcal{B}$.

To this end, let $x^{(1)}, \ldots, x^{(m)} \in [-1,1]$ denote univariate quadrature points, with associated quadrature weights $w^{(1)}, \ldots, w^{(m)} \in \R$. As we commented for the univariate basis functions $\phi_1, \ldots ,\phi_n \in L^2 ([-1,1])$, it is possible to choose a different set of quadrature points $x_k^{(1)},\ldots,x_k^{(m_k)}$ (and weights) for each dimension $k=1,\ldots ,d$, but for simplicity we simply adopt the same choice for each dimension and in particular $m_k = m$ for all $k=1,\ldots,d$.

To avoid confusion with the multi-index notation $\mathcal{I}$ that we use to denote basis coefficients, we will used $\mathcal{J} = ( j_1, \ldots, j_d ) $ to index the multivariate grid, defining the tensor $\mathcal{T}\in \mathbb{R}^{m \times \cdots \times m}$ of grid values for $\tilde{q}$ by 
\begin{equation}\label{eq:gridtensor}
\mathcal{T}[\mathcal{J}] := \tilde{q}(x^{(j_1)}, \ldots x^{(j_d)}).
\end{equation}
Now TT-cross can be applied precisely to determine a TT approximation $\mathcal{S} \approx \mathcal{T}$, only based on limited queries to the entries of the tensor $\mathcal{T}$. Let the cores of $\mathcal{S}$ be denoted $\mathcal{A}_1, \ldots ,\mathcal{A}_d$.

\begin{figure}[ht]
    \centering
    \setkeys{Gin}{draft=false}
    \includegraphics[width = 4in]{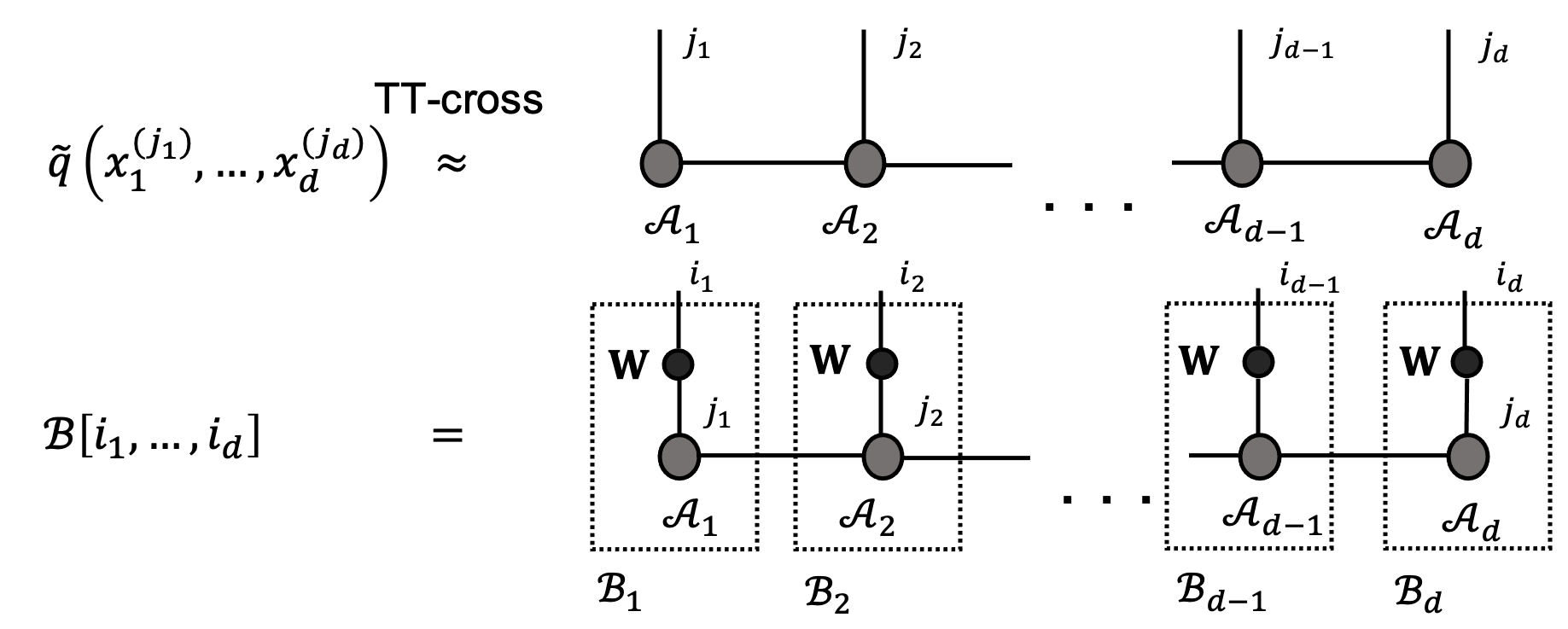}
    \caption{Computing the TT of basis coefficients $\mathcal{B}$. (a) depicts our tensor train $\mathcal{S}$ with cores $\mathcal{A}_1,\ldots,\mathcal{A}_d$, approximating the tensor $\mathcal{T}$ of the values of $\tilde{q}$ on our multivariate grid. (b) shows how $\mathcal{B}$ is constructed as a tensor train.
    }
    \label{tt-regression}
\end{figure}

In order to guarantee that $\tilde{q}_0 \approx \tilde{q}$, it suffices for the coefficient tensor to satisfy 
\begin{equation}\label{eq:Bapprox}
\mathcal{B}[\mathcal{I}] 
    \approx 
    \iprod{\tilde{q}(\mathbf{x}), \phi_{\mathcal{I}}(\mathbf{x})}_{L^2(\Omega)}.
\end{equation}
By replacing the integral on the right-hand side of \eqref{eq:Bapprox} with numerical quadrature and replacing the values of $\tilde{q}$ on the grid by the values furnished from the TT approximation $\mathcal{S}$, we can therefore construct $\mathcal{B}$ as in Figure~\ref{tt-regression}, where the weight matrix $\mathbf{W} \in \mathbb{R}^{n\times m}$ is defined by 
\[
\mathbf{W}[i,j] = \phi_i(x^{(j)}) w^{(j)}.
\]

Finally we discuss how the properly normalized $\mathcal{C}$ can be recovered from $\mathcal{B}$.
Recalling that $p_0 = q_0^2$, where
\[
{q}_{0}(\bx) = \sum_{\mathcal{I}} \mathcal{C}[\mathcal{I}],
\]
it can be seen readily using the orthonormality of the tensor product basis  that
\[
\int_\Omega p_0 \,d\bx = \Vert {q}_{0} \Vert_{L^2(\Omega)} 
= \Vert \mathcal{C} \Vert_\mathrm{F} = \langle \mathcal{C}[\mathcal{I}], \mathcal{C}[\mathcal{I}]\rangle_{\mathcal{I}}.
\]
In order to guarantee that our base distribution $p_0$ is in fact a normalized probability distribution, we therefore want to ensure $\Vert \mathcal{C} \Vert_\mathrm{F} = 1$.

In fact, $\Vert \mathcal{B} \Vert_\mathrm{F}$ can be computed by elementary TT contractions. Alternatively, if $\mathcal{B}$ is put into right-left orthogonal form, then $\Vert \mathcal{B} \Vert_\mathrm{F}$ can be recovered as the Frobenius norm of only the first tensor core. Therefore we are motivated to define
\begin{equation}\label{coefficient-tensor-C-is-just-B-normalized}
    \mathcal{C} = \frac{\mathcal{B}}{\norm{\mathcal{B}}_F},
\end{equation}
ensuring $\Vert \mathcal{C} \Vert_\mathrm{F} = 1$ as desired. The TT format for $\mathcal{C}$ can be recovered from that of $\mathcal{B}$ by rescaling only a single core, e.g., the first one (which would preserve right-left orthogonality, if $\mathcal{B}$ is already put into right-left orthogonal form).

\subsection{Flow model} \label{pushforwardtt} 
Following Section~\ref{vinf}, now that we have constructed a base distribution $p_0$, it remains only to discuss the choice of flow model $F_\theta^{(k)}$, as well as how to optimize the loss \eqref{eq:Ltheta_unpack}-\eqref{eq:Lhattheta_unpack}.

In this work we use flows based on invertible residual networks, or iResNets~\cite{https://doi.org/10.48550/arxiv.1811.00995,https://doi.org/10.48550/arxiv.1906.02735}. We choose the residual flow for its expressivity, as well as the convenience of initializing the flow map to be near the identity, as the base distribution may already offer a useful approximation of the target.

In the residual flow model, the flow maps are constructed as
\begin{equation}\label{residual-layer-has-a-neural-net}
    F_{\boldsymbol{\theta}}^{(k)} (\bx) = \bx + G_{\bt}^{(k)}(\bx),
\end{equation}
where the $G_\bt^{(k)}$ are themselves parametrized by feed-forward neural networks. In variational inference, we do not need to invert $F_{\boldsymbol{\theta}}^{(k)}$ computationally, though invertibility is guaranteed if $G_\bt^{(k)}$ is a contraction, and the Banach fixed point theorem enables efficient iterative computation of the inverse~\cite{784232}. 

However, as mentioned in Section~\ref{vinf}, it is important to be able to evaluate the Jacobian determinant of $F_\bt^{(k)}$. This can be achieved by Taylor series expansion:
\begin{equation}\label{eq:matrix-trace-formula}
    \log \abs{DF_\bt^{(k)}} = \log \abs{I_d + DG_\bt^{(k)}} =
    \Tr(\log( I_d + DG_\bt^{(k)} )) 
    = \sum_{m=1}^{\infty}(-1)^{m+1}\frac{\Tr[ (DG_\bt^{(k)})^m ]}{m}.
\end{equation}
We can truncate the series (\ref{eq:matrix-trace-formula}) and use stochastic trace approximation via Hutchinson's estimator~\cite{9400306} to accelerate the computation.

To accelerate training, we opted to add batch normalization \cite{https://doi.org/10.48550/arxiv.1502.03167} following each residual connection. Furthermore, since our base distribution is already rather close to the target distribution, we want to initialize the weights such that the map is near identity. To this end, all weights in the residual connection $g_{\boldsymbol{\theta}}$ are generated from a uniform distribution on $[-0.25, 0.25]$.

\subsection{Summary}\label{main-routine}
Here we summarize our proposed method for variational inference on the density $p(\bx) \propto \exp(-U(\bx))$, given the ability to evaluate the function $U(\bx)$.
\begin{enumerate}
    \item Use TT-cross to construct a tensor train $\mathcal{S}$ approximating the tensor $\mathcal{T}$~\eqref{eq:gridtensor} of values of $\tilde{q}(\bx) := \exp(-U(\bx)/2)$ on our multivariate grid.
    \item Via the construction of Figure~\ref{tt-regression} and~\eqref{coefficient-tensor-C-is-just-B-normalized}, 
    build a tensor train $\mathcal{C}$ of basis coefficients, specifying $q_0(\bx)$ via~\eqref{tensor-representation}. By construction, $p_0(\bx) := q_0(\bx)^2$ is normalized and approximates the target $p(\bx)$.
    \item Draw a corpus of samples $\bz^{(1)}, \ldots , \bz^{(S)} \sim p_0$, equipped with densities $p_0(\bz^{(s)})$, $s=1,\ldots,S$, following the algorithm described in Section~\ref{sec:sampling}.
    \item Optimize the empirical loss \eqref{eq:Lhattheta_unpack}, using the flow model specified in Section~\ref{pushforwardtt}.
\end{enumerate}
Further details of the training will be discussed in the Section~\ref{numerical-experiments} below, where numerical experiments are reviewed.

\section{Numerical experiments}\label{numerical-experiments}
In this section, we demonstrate the ability of tensorizing flow to learn high-dimensional target distributions. We describe implementation details for constructing the TT base distribution in Section~\ref{tensor-train-hyperparams} and for flow training in Section~\ref{flowtable}. In Section~\ref{examples-section}, we present experiments on a Gaussian mixture distribution (Section~\ref{mixture-gaussian-section}),
and Ginzburg-Landau models (Sections~\ref{gl1d-section} and \ref{gl2d-section}).

\subsection{Details of base distribution construction}\label{tensor-train-hyperparams}

Recall that in our construction of the base distribution, we apply TT-cross to the tensor $\mathcal{T}$ defined in \eqref{eq:gridtensor} to obtain a tensor train $\mathcal{S}$. For our various experiments, we record in Table~\ref{ranktable} the maximum TT rank of $\mathcal{S}$, which is also the maximum TT rank of the coefficient tensor $\mathcal{C}$. For benchmarking purposes, we also obtain a ground truth coefficient tensor $\mathcal{C}_{\text{true}}$ by specifying a local relative accuracy of $10^{-10}$ in TT-cross and allowing the ranks to be as large as needed to ensure this. These ranks are generally much higher than those of the low-rank approximations we use in our experiments and are also reported in Table~\ref{ranktable}, along with the number $n$ of univariate basis functions $\phi_1 ,\ldots, \phi_n$, which we choose to be the Legendre polynomials.

\vspace{2mm}

\begin{center}
\begin{tabular}{ |c|c|c|c|c| } 
 \hline
 Experiment & $d$ & $n $  & Max rank of $\mathcal{C}_{\text{true}}$ & Max rank of $\mathcal{C}$ \\
 \hline
 Gaussian mixture      & 30 &   512  & 116  & 2    \\
  1-D Ginzburg-Landau                 & 25 &   50   & 30   & 2       \\ 		
  2-D Ginzburg-Landau                 & 64 &   30   & 694  & 3     \\ 		
 \hline
\end{tabular}
\captionof{table}{Details of base distribution construction for each experiment.}\label{ranktable}
\end{center} 

Computationally, we shift and scale all of our problems to the cube $[-1,1]^d$, and we initialize the base distribution of all normalizing flows as $\mathcal{N}(0,0.2 I)$, chosen to ensure that all samples fall within the computational domain.  In reporting our results, we shift and scale back to the original bounding box for the support of $p(\bx)$.

\subsection{Details of flow model} We use a multi-layer perceptron (MLP) neural-network architecture with ReLU activations to define the residual connections in (\ref{residual-layer-has-a-neural-net}). The numbers of hidden layers (network depths) and layer widths used in our experiments are reported in Table~\ref{residual-flow-params}. For training, we optimize with Adam~\cite{https://doi.org/10.48550/arxiv.1412.6980}, along with an exponential learning rate scheduler with a decay factor of 0.9999. We summarize the training hyperparameters in Table \ref{flowtable}. As observed in \cite{https://doi.org/10.48550/arxiv.2108.12657}, a smaller learning rate and larger batch size tend to yield stabler convergence of the model. To further stabilize the training, we implement gradient clipping such that the entries of the gradient always lie in the range $[-10^4, 10^4]$.

\vspace{2mm}

\begin{center}
\begin{tabular}{ |c|c|c| } 
 \hline
 Experiment & Layer width &  Network depth  \\ 
 \hline
  Gaussian mixture     &  32    &  5    \\
  1-D Ginzburg-Landau    &  32    &  5    \\ 		
  2-D Ginzburg-Landau    &  64    &  5    \\ 		
 \hline
\end{tabular}
\captionof{table}{Neural network widths and depths.}\label{residual-flow-params}
\end{center}

\begin{center}
\begin{tabular}{ |c|c|c|c|c| } 
 \hline
 Experiment & Batch size & Learning rate  & Number of epochs & Flow length \\ 
 \hline
 Gaussian mixture       & 128 &   $5\times 10^{-4}$ & 200 & 10 \\
  1-D Ginzburg-Landau                & 256 &   $5\times 10^{-4}$ & 200 & 12 \\ 	
  2-D Ginzburg-Landau                 & 64 &   $3\times 10^{-5}$  & 200 & 12 \\
 \hline
\end{tabular}
\captionof{table}{Training hyperparameters.}\label{flowtable}
\end{center}

For consistency, we compare the performance of TF and NF under the exact same training conditions. All flow models are trained using a training set of size $S_{\text{train}} = 10^4$ and validated on a holdout set of size $S_{\text{holdout}} = 10^4$. For both TF and NF, we will plot the loss \eqref{eq:Lhattheta_unpack} evaluated using the holdout samples. The validation loss will be reported as an average over 10 training runs with random network initializations.

\subsection{Examples}\label{examples-section}
Here we apply the {tensorizing flow} and compare its performance with traditional normalizing flow on a range of high-dimensional densities.

To quantify the performance of TF relative to that of NF, we define $-\log Z_{\text{true}}$ to be the loss~\eqref{eq:Lhattheta_unpack} estimated from a very high-fidelity tensor train, obtained as described in Section~\ref{tensor-train-hyperparams}. Likewise we define $-\log Z_{\text{TF}}$ and $-\log Z_{\text{NF}}$ to be the final losses after training of the tensorizing and normalizing flows, respectively. Then we define the \emph{error ratio}:
\begin{equation}\label{defining-error-ratio}
    \text{Error Ratio} := \frac{{\log Z_{\text{true}} - \log Z_{\text{TF}}}}{{\log Z_{\text{true}} - \log Z_{\text{NF}}}}.
\end{equation}
Note that both the numerator and denominator must be positive.

\subsubsection{\normalfont\emph{Gaussian mixture model}}\label{mixture-gaussian-section} First we consider a synthetic example in dimension $d=30$, defined as a mixture of five Gaussians. The means are:
$
	\boldsymbol{\mu}^{(1)} = [0, \cdots, 0, 2, 2]^\top
$, 
$
\boldsymbol{\mu}^{(2)} = 
	[0, \cdots, 0, 2, -2]^\top 
$, 
$\boldsymbol{\mu}^{(3)} = 
	[0, \cdots, 0, -2, 2]^\top
$,
$\boldsymbol{\mu}^{(4)} = [0, \cdots, 0, -2, -2]^\top 
$, and 
$ \boldsymbol{\mu}^{(5)} = [0, \cdots, 0, 0, 0]^\top.
$
Then we define matrices $\boldsymbol{\Sigma}^{(i)}$, $i=1,\ldots,5$ as the identity, except in the $2\times 2 $ block corresponding to the last two dimensions, where we take:
$$
	\boldsymbol{\Sigma}^{(1)}_{d-1:d, d-1:d} = \boldsymbol{\Sigma}^{(4)}_{d-1:d, d-1:d} = 
	\begin{bmatrix}
		1 &  {19}/{20}\\
		 {19}/{20} & 1
	\end{bmatrix}
$$
and 
$$
	\boldsymbol{\Sigma}^{(2)}_{d-1:d, d-1:d} = \boldsymbol{\Sigma}^{(3)}_{d-1:d, d-1:d} = 
	\begin{bmatrix}
		1 &  -{19}/{20}\\
		 -{19}/{20} & 1
	\end{bmatrix}.
$$ 
Then we define a $p(\bx)$ as the equally weighted mixture of the Gaussian distributions $\mathcal{N}(\boldsymbol{\mu}^{(i)}, 0.4 \times \boldsymbol{\Sigma}^{(i)})$, $i=1,\ldots, 5$.

\begin{figure}[!htb]
\centering
\setkeys{Gin}{draft=false}
\begin{tabular}{ccc}
Target & TF & NF \\
\subfloat[$(x_{28},x_{29})$]{\includegraphics[width = 1.43in]{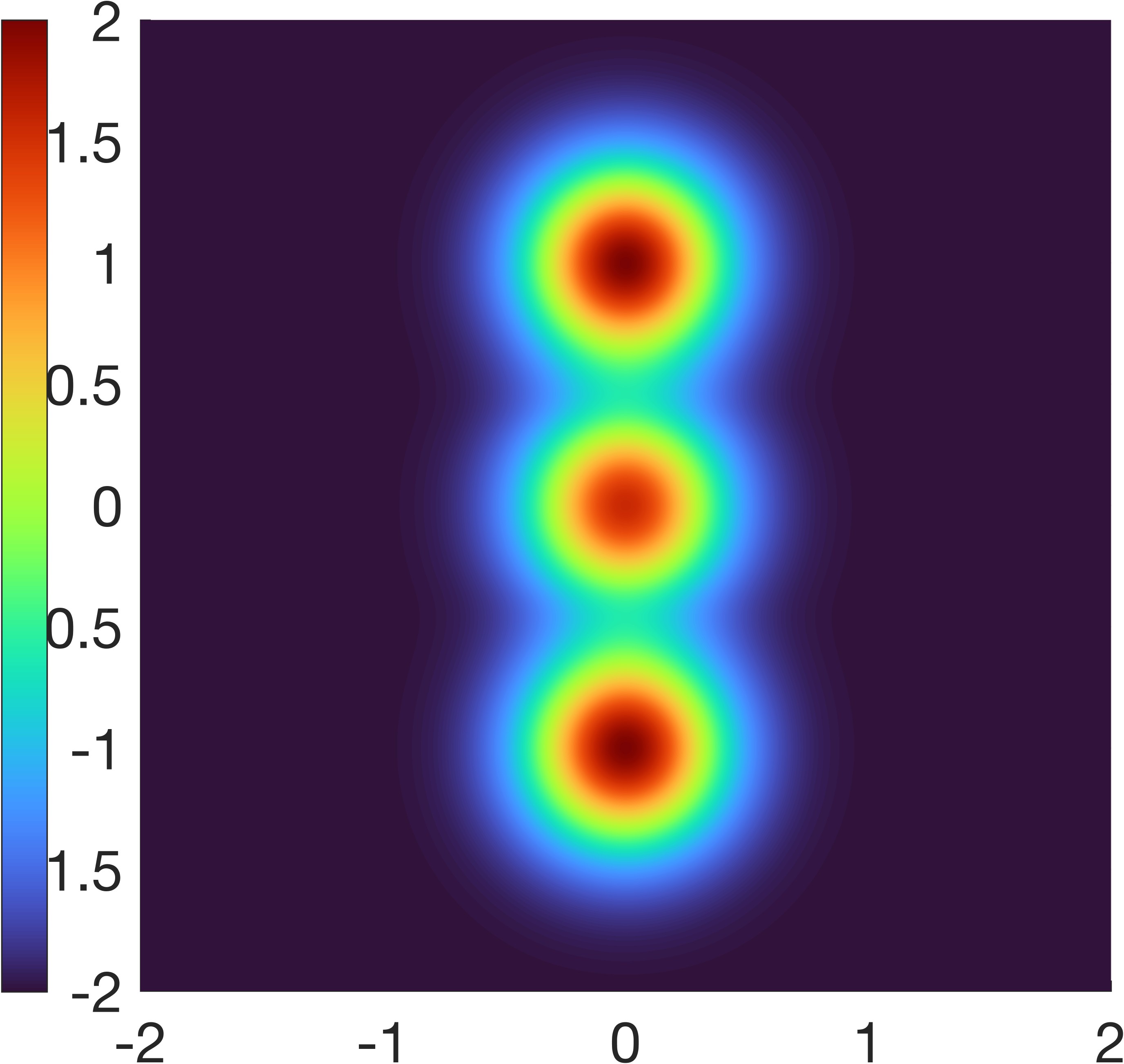} \label{mixture-true1} } &
\subfloat[$(x_{28},x_{29})$]{\includegraphics[width = 1.45in]{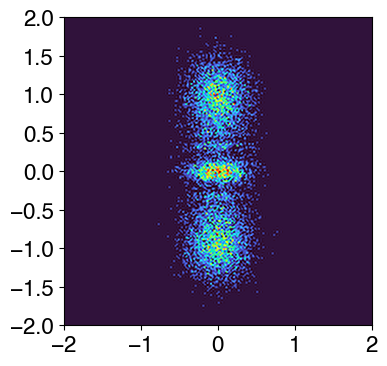} \label{mixture-tf1}} &
\subfloat[$(x_{28},x_{29})$]{\includegraphics[width = 1.45in]{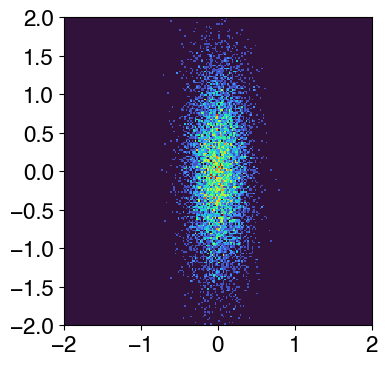} \label{mixture-nf1}}  \\
\subfloat[$(x_{29},x_{30})$]{\includegraphics[width = 1.43in]{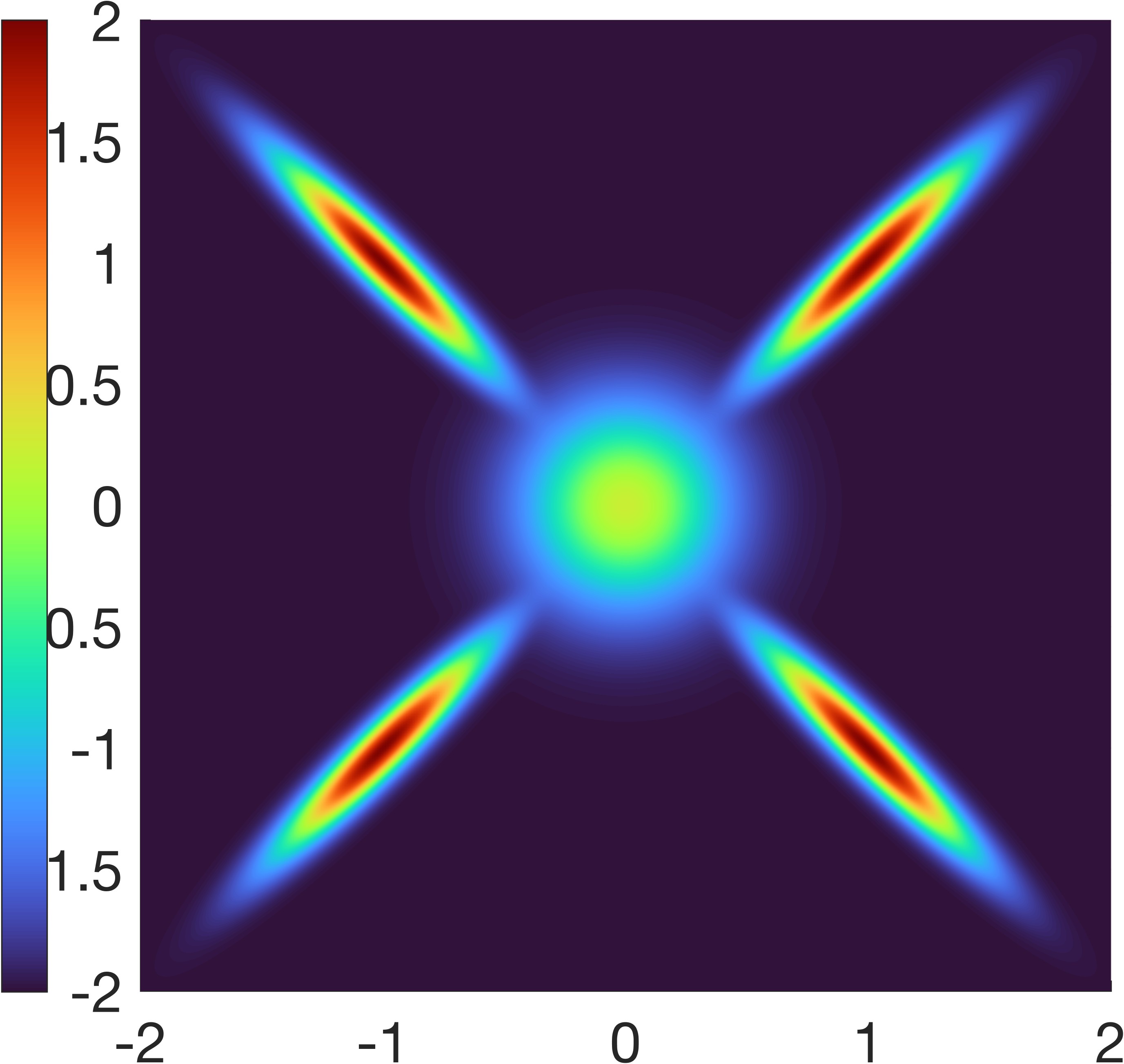} \label{mixture-true2}} &
\subfloat[$(x_{29},x_{30})$]{\includegraphics[width = 1.45in]{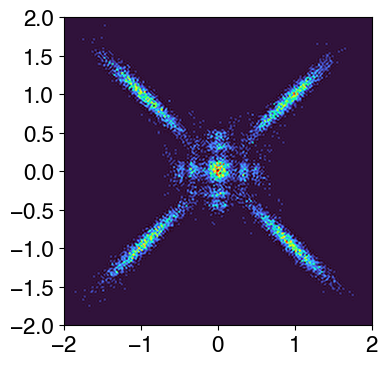} \label{mixture-tf2}} &
\subfloat[$(x_{29},x_{30})$]{\includegraphics[width = 1.45in]{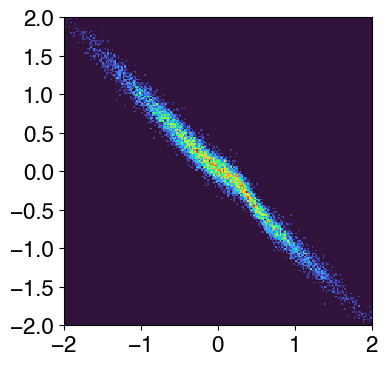} \label{mixture-nf2}}
\end{tabular}
\caption{Marginals of the Gaussian mixture model corresponding to variables $(x_{28},x_{29})$ and  $(x_{29},x_{30})$. (a)(d): Ground truth marginal densities. (b)(e): Marginal density estimates obtained from sampling the trained TF. (c)(f): Marginal density estimates obtained from sampling the trained NF.}
\label{fig:mixture}
\end{figure}

Figures~\ref{fig:mixture} (a) and (d) depict various marginals of $p$, demonstrating multimodality clearly. Figure~\ref{fig:mixture} (b), (c), (e), and (f) depict marginal density estimates produced from the trained NF and TF. 

\begin{figure}[!htb]
    \centering
    \setkeys{Gin}{draft=false}
    \begin{tabular}{cc}
         \subfloat[TT Base: $(x_{28},x_{29})$]{\includegraphics[width = 1.5in]{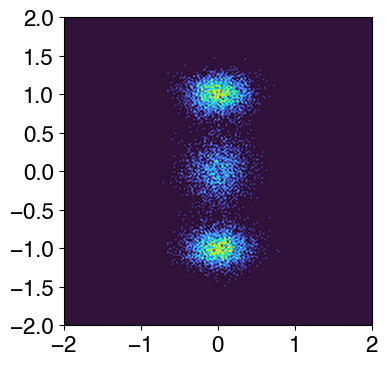} \label{mixture-before1} } 
         &
         \subfloat[$(x_{29},x_{30})$]{\includegraphics[width = 1.5in]{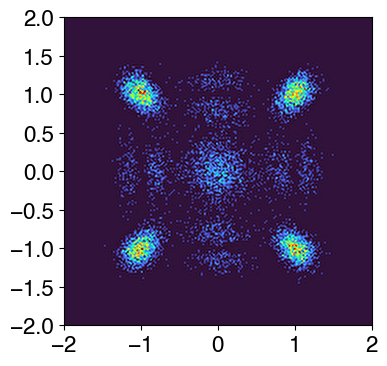} \label{mixture-before2} }
    \end{tabular}
    \caption{Estimated marginals of the Gaussian mixture model obtained directly from the rank-2 tensor train base distribution $p_0$, without flow.}
    \label{mixture-before}
\end{figure}

For comparison, corresponding marginal density estimates produced from the TT base distribution $p_0$ are presented in Figure~\ref{mixture-before}.

\begin{figure}[!htb]
\centering
\setkeys{Gin}{draft=false}
\includegraphics[width=3in]{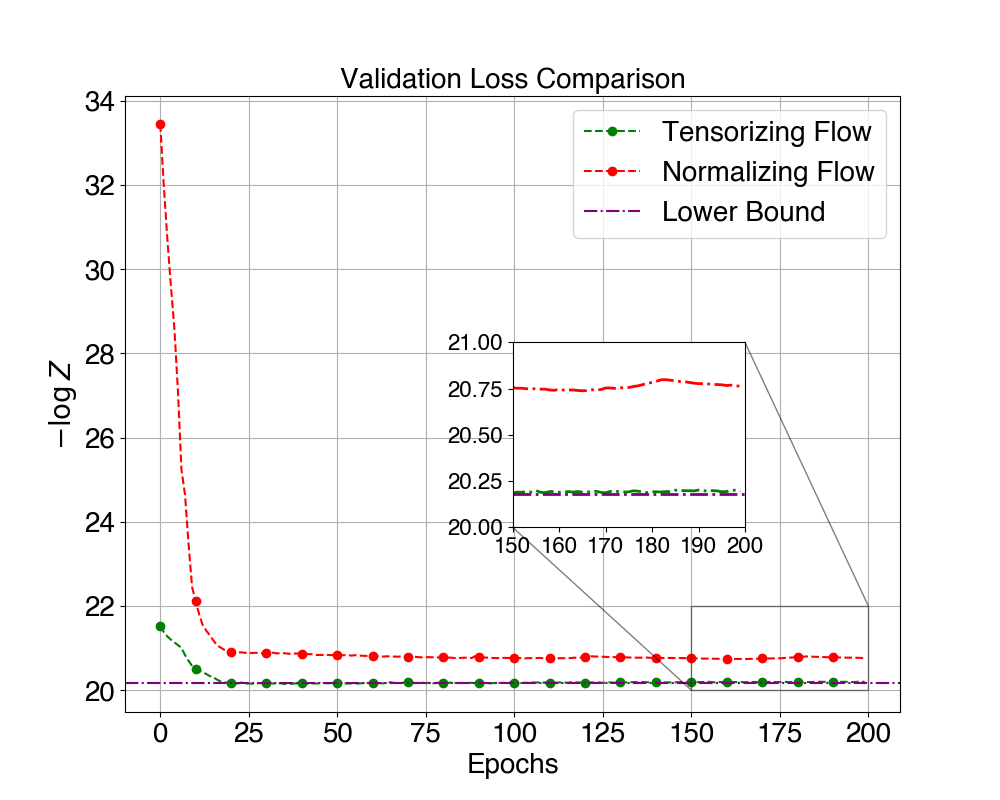}
\caption{Validation loss plot for TF and NF over the course of training for the Gaussian mixture distribution.}\label{mixture-gaussian-loss-profile}
\end{figure}

We can see that the TF is more capable of capturing multimodality of the target, even given a base distribution that is very coarse, involving only a TT of rank 2, which is the minimal nontrivial rank. Meanwhile the NF fails to capture the modes with a unimodal base distribution. We quantitatively confirm the effectiveness of TF in the validation loss plot of Figure~\ref{mixture-gaussian-loss-profile} and Table~\ref{mixturedetails}.

\
\newline
\begin{centering}
\begin{tabular}{ |c|c|c|c|c| } 
 \hline
 Base distribution & Start & End & $-\log Z_{\text{true}}$   & Error Ratio \\ 
 \hline
 TT (rank 2) & 21.5266 & 20.1952 & \multirow{2}{*}{\underline{20.1794}} &  
 \multirow{2}{*}{\underline{0.0271}}
 \\
  Gaussian & 33.4478 & 20.7634 &        & \\ 		
 \hline
\end{tabular}
\captionof{table}{Training synopsis for TF and NF applied to the Gaussian mixture model.}\label{mixturedetails}
\end{centering}

\subsubsection{\normalfont\emph{One-dimensional Ginzburg-Landau model}}\label{gl1d-section} Ginzburg-Landau theory offers a variety of nontrivial sampling problems relevant to the study of phase transitions and rare events~\cite{Huebener1979}. Traditional sampling methods are often inadequate even for simple models \cite{doi:10.1063/1.5110439}. In this section, we consider a Ginzburg-Landau-type model for an antiferromagnet on a 1D lattice, where the states of neighboring sites prefer to orient oppositely. More precisely, we denote the states at $(d+2)$ lattice points as $u_0,u_1,\ldots,u_{d+1} \in \mathbb{R}$, subject to the boundary conditions $u_0=u_{d+1}=0$. The energy function is defined as follows:
\begin{equation}\label{gl1d-energy}
	E(u_1,\ldots, u_{d}) := 
	-\sum_{i=1}^{d+1}\frac{\delta}{2}\bigg(\frac{u_{i}-u_{i-1}}{h}\bigg)^2 + \frac{1}{4\delta} \sum_{i=1}^d (1-u_i^2)^2,
\end{equation}
where $\delta >0 $ is a scalar parameter and $h > 0$ is the lattice spacing. The final term in (\ref{gl1d-energy}) is a double-well potential ensuring (when $\delta$ is small) that the $u_i$ take values near $\pm 1$. The target probability density is then defined
\begin{equation}\label{boltzmann-gibbs-form}
    p(\boldsymbol{u}) \propto \exp(-\beta E(\boldsymbol{u})),
\end{equation}
where $\beta$ is the inverse temperature. The energy function has two minimizers, yielding a bimodal distribution.

\begin{figure}[htb]
\centering
\setkeys{Gin}{draft=false}
\includegraphics[width=2.5in]{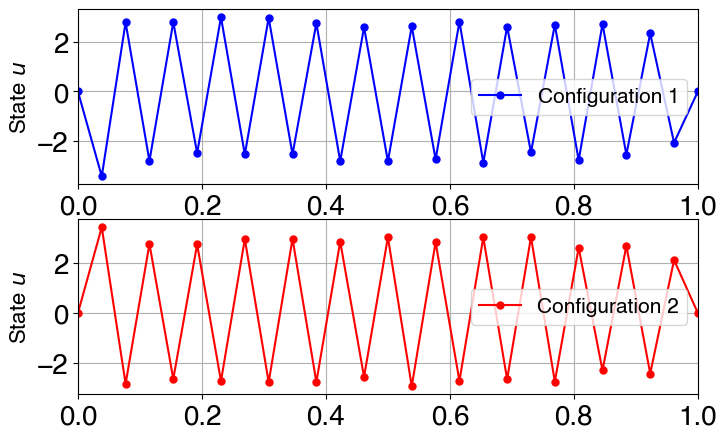}
\caption{One typical sample configuration from each mode of \eqref{boltzmann-gibbs-form}.}
\label{gl1d-bimodal}
\end{figure}

In our experiment, we set $d=35$, $h=1/(d+1)=1/36$, $\beta=6.25\times 10^{-2}$, and $\delta=0.04$.
We visualize one typical representative of each mode (drawn from $p_\mathrm{true}$) in Figure~\ref{gl1d-bimodal}.

\begin{figure}[htb]
\centering
\setkeys{Gin}{draft=false}
\begin{tabular}{c}
\begin{tabular}{l}
\subfloat[Base]{\includegraphics[width=4.5in]{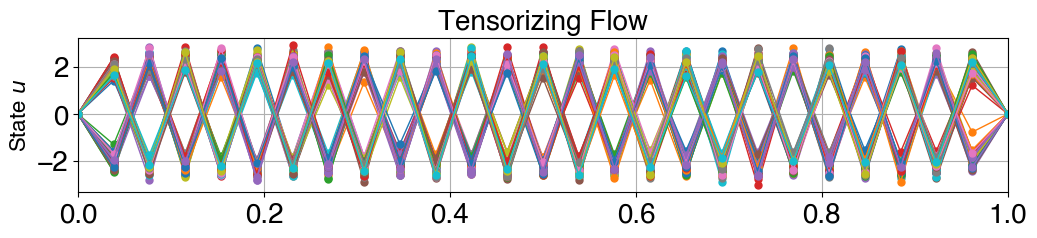} \label{gl1d-tfbase} }
\end{tabular} \\
\begin{tabular}{l}
\subfloat[pushforward]{\includegraphics[width=4.5in]{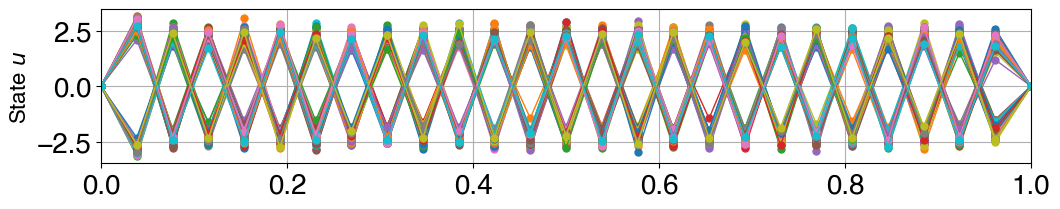} \label{gl1dtfpush} }
\end{tabular}
\end{tabular} % end main table
\caption{Several typical sample configurations of the 1-dimensional Ginzburg-Landau model drawn from the TF (a) before training and (b) after training.} \label{gl1d-result-tf}
\end{figure}

\begin{figure}[htb]
\centering
\setkeys{Gin}{draft=false}
\begin{tabular}{c}
\begin{tabular}{l}
\subfloat[Base]{\includegraphics[width=4.5in]{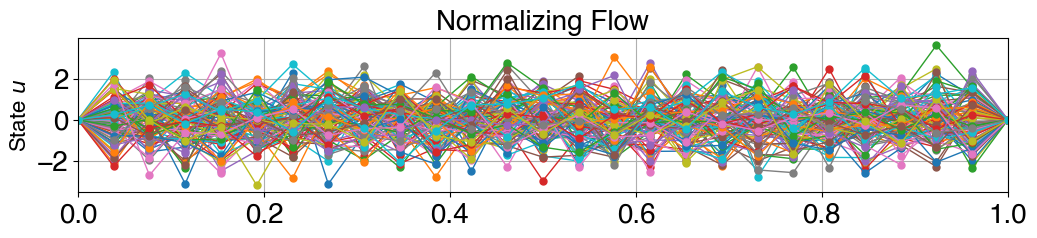}\label{gl1dnfbase}}
\end{tabular} 
\\
\begin{tabular}{l}
\subfloat[pushforward]{\includegraphics[width=4.5in]{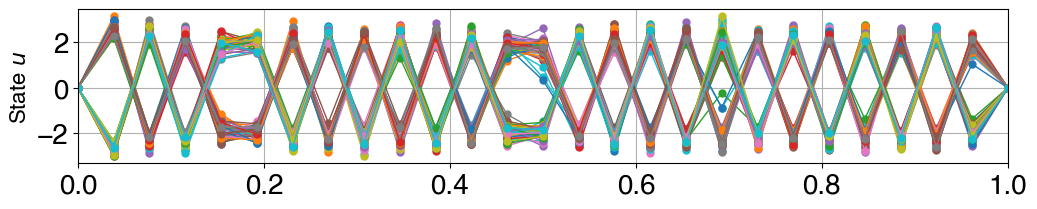}\label{gl1dnfpush}}
\end{tabular}
\end{tabular} % end main table
\caption{Several typical sample configurations of the 1-dimensional Ginzburg-Landau model drawn from the NF (a) before training and (b) after training.} \label{gl1d-result-nf}
\end{figure}

\begin{figure}[!htb]
\centering
\setkeys{Gin}{draft=false}
\includegraphics[width=2.5in]{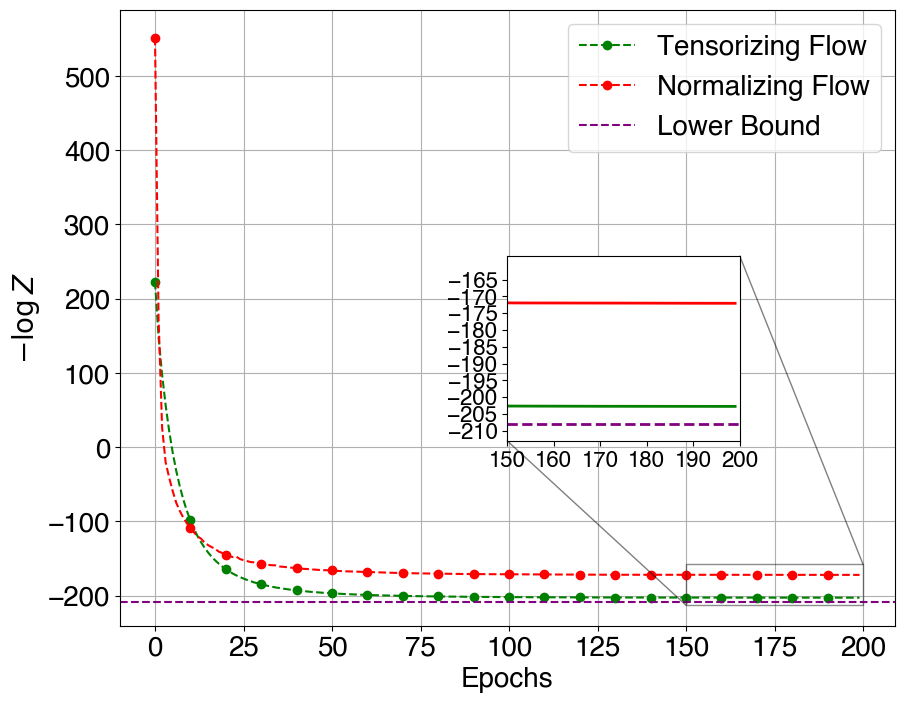}
\caption{Validation loss plot for TF and NF over the course of training for the 1-dimensional Ginzburg-Landau model.}\label{gl1d-loss}
\end{figure}

Typical sample configurations generated by the TF and NF, before and after training, are plotted in Figures~\ref{gl1d-result-tf} and~\ref{gl1d-result-nf}, respectively. We observe that the trained NF samples sometimes appear to be qualitatively incorrect, and the improvement of TF over NF is confirmed quantitatively in terms of the validation loss, as summarized in Table~\ref{gl1d-details} and plotted in Figure~\ref{gl1d-loss}.

\vspace{4mm}
\begin{centering}
\begin{tabular}{ |c|c|c|c|c| } 
 \hline
 Base Distribution & Start & End & $-\log Z_{\text{true}}$ & Error Ratio \\ 
 \hline
 TT (rank 2) & 222.736 & $-202.756$ & \multirow{2}{*}{\underline{$-208.014$}} &  
 \multirow{2}{*}{\underline{0.146}}
 \\
  Gaussian & 551.411 & $-172.069$ &        &  \\ 		
 \hline
\end{tabular}
\captionof{table}{Training synopsis for TF and NF applied to the 1-dimensional Ginzburg-Landua model.}\label{gl1d-details}
\end{centering}

\subsubsection{\normalfont\emph{Two-Dimensional Ginzburg-Landau model}}\label{gl2d-section}
In this section, we consider an analogous Ginzburg-Landau-type model on a two-dimensional, $(d+2)\times (d+2) $ square lattice. We denote the values of the scalar field at lattice points as $u_{i,j}$, $i,j=0,\ldots,d+1$, and define the energy:
\begin{equation}\label{gl2d-energy}
	E(\boldsymbol{u}) := 
	\frac{\delta}{2} \bigg[
	\sum_{i=1}^{d+1}\sum_{j=1}^{d+1}
		\bigg(\frac{u_{i,j} - u_{i-1,j}}{h}\bigg)^2 
		+
		\bigg(\frac{u_{i,j} - u_{i,j-1}}{h}\bigg)^2 
	\bigg] + 
	\frac{1}{4\delta}
	\sum_{i=1}^{d}\sum_{j=1}^{d}
	(1-u_{i,j}^2)^2,
\end{equation}
with boundary conditions 
\begin{equation}\label{gl2d-boundary-condition}
    u_{0,:}=u_{d+1,:} = \mathbf{1}, \quad 
    u_{:,0}=u_{:,d} = -\mathbf{1}.
\end{equation}
Once again $\delta > 0$ is a scalar parameter, and $h>0$ is the lattice spacing. The target distribution is again defined by 
\begin{equation}\label{gl2d-density}
    p(\boldsymbol{u}) \propto \exp(-\beta E(\boldsymbol{u})).
\end{equation}
We set $d=64$, $h=1/(d+1)=1/9$, $\beta=0.2$, and $\delta=0.04$. 

\begin{figure}[!htb]
\centering
\setkeys{Gin}{draft=false}
\begin{tabular}{cc}
TF & NF \\
\subfloat[]{\includegraphics[width=2.2in]{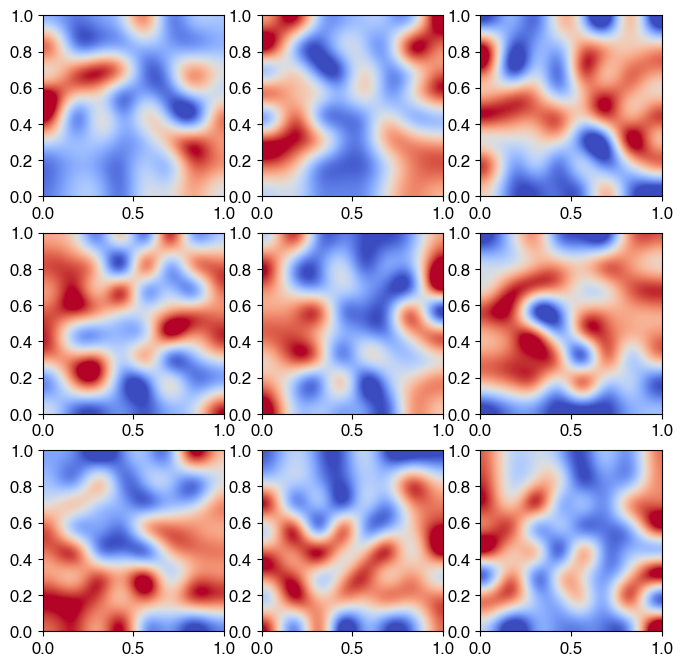} \label{gl2dtfsamples}} &
\subfloat[]{\includegraphics[width=2.2in]{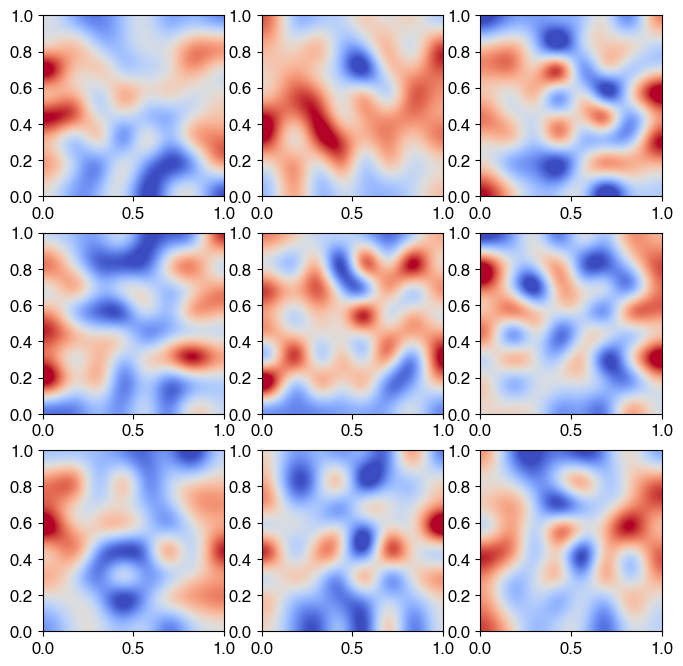} \label{gl2dnfsamples}}
\end{tabular}
\caption{Several typical sample configurations of the 2-dimensional Ginzburg-Landau model drawn from (a) the trained TF and (b) the trained NF.}\label{random_gl2d_states}
\end{figure}

Typical sample configurations generated by the trained TF and NF are plotted in Figure~\ref{random_gl2d_states}. Qualitative differences are visually apparent.

\begin{figure}[htb]
\centering
\setkeys{Gin}{draft=false}
\begin{tabular}{ccc}
\subfloat[$\overline{u}_{\text{true}}$]{\includegraphics[width=4cm]{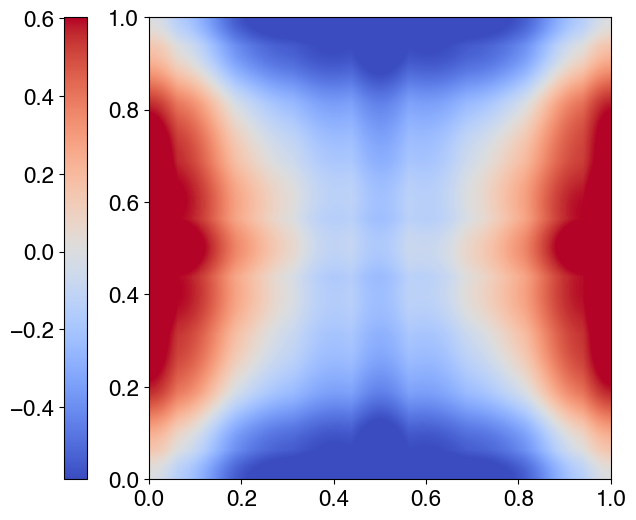} \label{gl2d-target-average} } &
\subfloat[$\overline{u}_{\text{TF}}$]{\includegraphics[width=4cm]{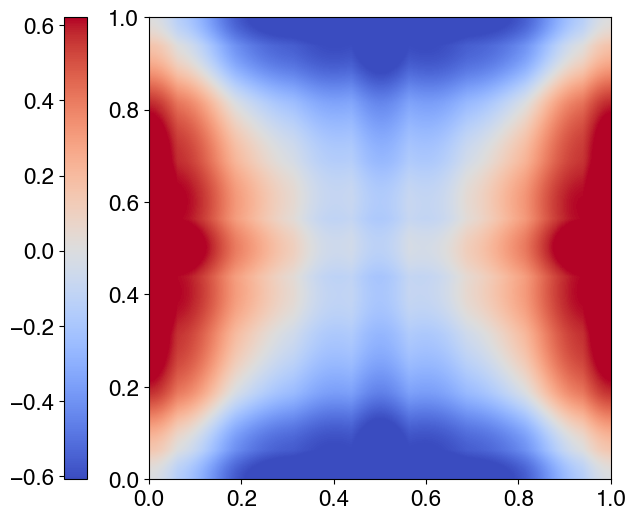}} &
\subfloat[$\overline{u}_{\text{NF}}$]{\includegraphics[width=4cm]{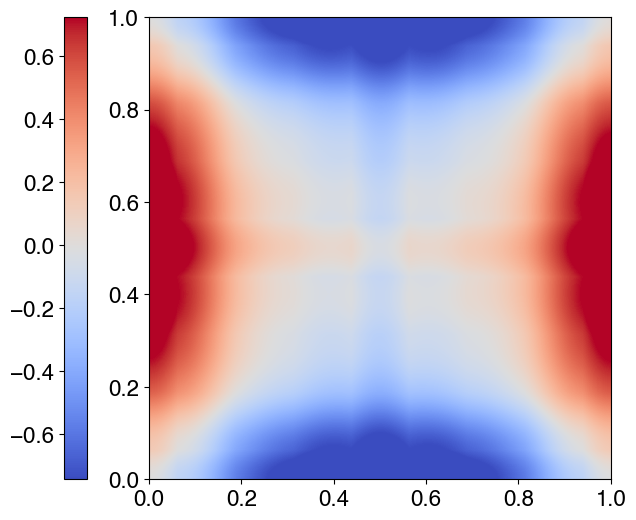}} 
\end{tabular}
\caption{GL 2-Dim: Estimated first moment of sampled states $\overline{u}$ defined in (\ref{sample-average-states}). (a) Ground truth (b) TF (c) NF. All $\overline{u}_{\text{true}}$, $\overline{u}_{\text{TF}}$, and $\overline{u}_{\text{NF}}$ are computed with $S = 20000$ samples. } \label{gl2d-average-surface}
\end{figure}

\begin{figure}[htb]
\centering
\setkeys{Gin}{draft=false}
\begin{tabular}{cc}
TF & NF \\
\subfloat[$\abs{\overline{u}_{\text{TF}} - \overline{u}_{\text{true}}}$]{\includegraphics[width=6cm]{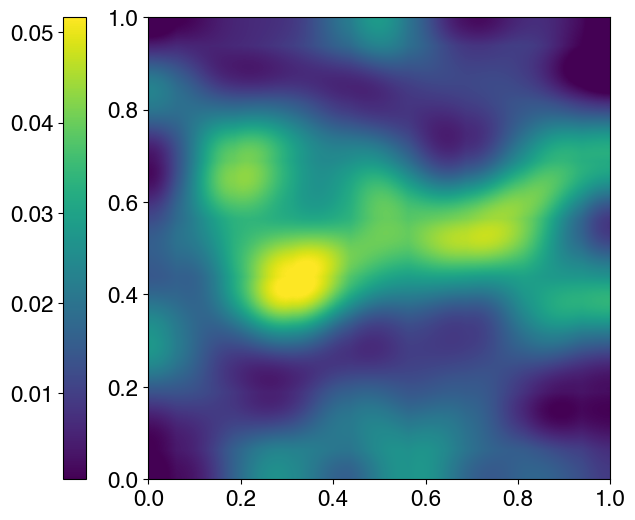}} &
\subfloat[$\abs{\overline{u}_{\text{NF}} - \overline{u}_{\text{true}}}$]{\includegraphics[width=6cm]{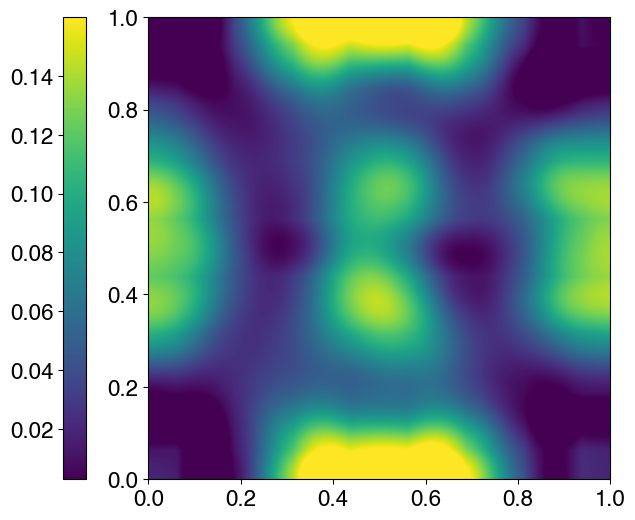}}
\end{tabular}
\caption{GL 2-Dim: (a) Pointwise absolute errors between $\overline{u}_{\text{TF}}$ and $\overline{u}_{\text{true}}$, defined in (\ref{sample-average-states}). (b) Pointwise absolute errors between $\overline{u}_{\text{NF}}$ and $\overline{u}_{\text{true}}$, defined in (\ref{sample-average-states}). Errors are plotted with respect to each lattice point $(i,j)$ in $[0,1]^2$. }\label{gl2d-average-error}
\end{figure}

In Figure~\ref{gl2d-average-surface}, we plot empirical estimates for:
\begin{equation}\label{sample-average-states}
    \overline{u}_{i,j} := \mathbb{E} [u_{i,j}]
\end{equation}
obtained from $S = 2 \times 10^4$ samples from both the TF and NF, as well as the reference $p_{\text{true}}$, which we denote respectively as $\overline{u}_{\text{TF}}$, $\overline{u}_{\text{NF}}$, and $\overline{u}_{\text{true}}$. In Figure~\ref{gl2d-average-error}, we plot the absolute errors of $\overline{u}_{\text{TF}}$ and $\overline{u}_{\text{NF}}$, demonstrating the advantage of TF more clearly.

\begin{figure}[H]
	\centering
	\setkeys{Gin}{draft=false}
	\includegraphics[width=7cm]{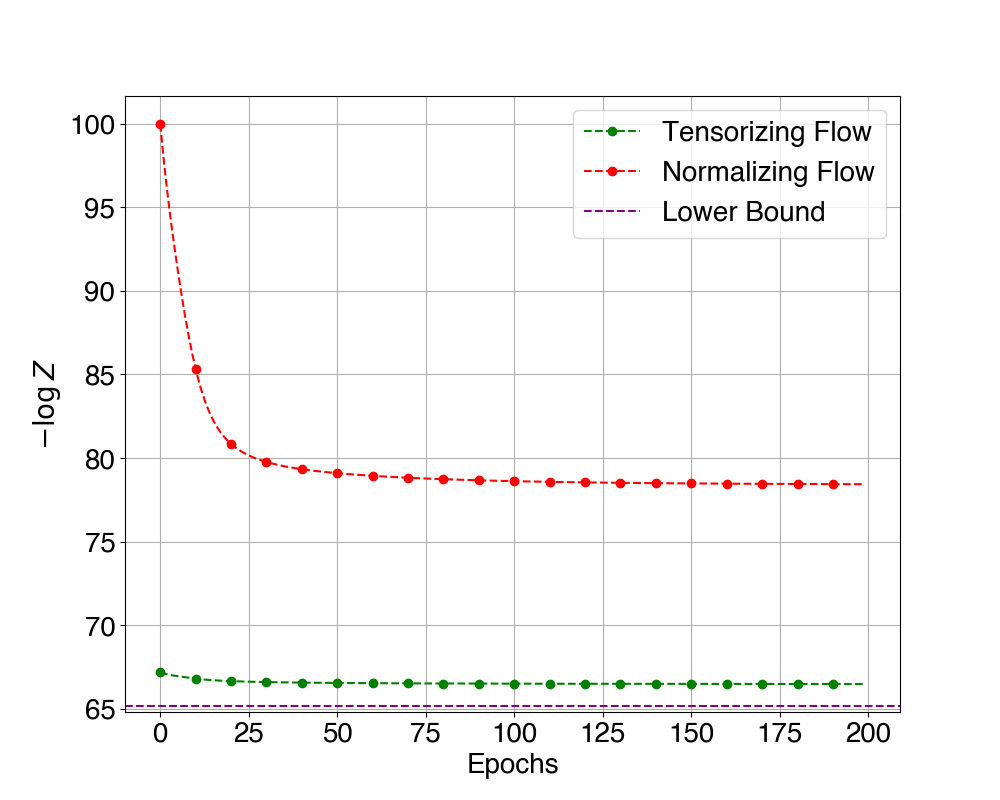}
	\caption{Validation loss plot for TF and NF over the course of training for the 2-dimensional Ginzburg-Landau model.}\label{gl2d-loss-comparison}
\end{figure}

The improvement of TF over NF is again confirmed quantitatively via the validation loss, plotted in Figure~\ref{gl2d-loss-comparison} and summarized in Table~\ref{gl2d-details}.

\vspace{2mm}
\begin{centering}
\begin{tabular}{ |c|c|c|c|c| } 
\hline
Base Distribution & Start & End & $-\log Z_{\text{true}}$ & Error Ratio \\
\hline
 Rank 3 TT & 67.202 & 66.488 & \multirow{2}{*}{\underline{65.198}} &  
\multirow{2}{*}{\underline{0.0974}}
\\
$\mathcal{N}(\mathbf{0},0.2 \mathbf{I}_{64})$ & 99.979 & 78.438 & &  \\ 
\hline
\end{tabular}
\captionof{table}{Training synopsis for TF and NF applied to the 2-dimensional Ginzburg-Landua model.}\label{gl2d-details}
\end{centering}

\section{Conclusion}

In this paper, we propose a flexible ansatz for variational inference, \emph{tensorizing-flow}, by combining a neural-network flow model with a low rank TT base distribution. Such a base distribution can be obtained via numerical linear algebra routines and captures non-trivial overlaps with the target distribution. The representation power of the neural-network flow model then serves to bridge the approximation gap. Finally, effectiveness of the TT base measure is concretized by tighter variational lower bound convergence in several high-dimensional examples.

\bibliographystyle{siam}
\bibliography{ref}

\end{document}